\pdfoutput=1

\documentclass[11pt]{article}
\usepackage[preprint]{acl}

\usepackage{times}
\usepackage{latexsym}

\usepackage[T1]{fontenc}

\usepackage[utf8]{inputenc}

\usepackage{microtype}

\usepackage{inconsolata}

\usepackage{graphicx}
\usepackage{xcolor}
\definecolor{darkgreen}{RGB}{0,100,0}
\usepackage{listings}
\usepackage{booktabs}
\title{SDE-SQL: Enhancing Text-to-SQL Generation in Large Language Models via Self-Driven Exploration with SQL Probes}

\author{
\textbf{Wenxuan Xie\textsuperscript{1}},
\textbf{Yaxun Dai\textsuperscript{2}},
\textbf{Wenhao Jiang\textsuperscript{3}\thanks{Corresponding author}} \\
\textsuperscript{1}South China University of Technology, 
\textsuperscript{2}Soochow University\\
\textsuperscript{3}Guangdong Laboratory of AI and Digital Economy (SZ) \\
\small{
\href{mailto:lancelotxie601@gmail.com}{lancelotxie601@gmail.com}, 
\href{mailto:cswhjiang@gmail.com}{cswhjiang@gmail.com}
}
}

\begin{document}
\maketitle
\begin{abstract}
Recent advances in large language models (LLMs) have led to substantial progress on the Text-to-SQL task. However, existing approaches typically depend on static, pre-processed database information supplied at inference time, which restricts the model’s capacity to deeply comprehend the underlying database content. In the absence of dynamic interaction, LLMs are limited to fixed, human-curated context and lack the ability to autonomously query or explore the data. To overcome this limitation, we introduce \textbf{SDE-SQL}, a novel framework that empowers LLMs to perform \textbf{Self-Driven Exploration} of databases during inference. This is achieved through the generation and execution of \textbf{SQL probes}, enabling the model to actively retrieve information and iteratively refine its understanding of the database. Unlike prior methods, \textbf{SDE-SQL} operates in a \textbf{zero-shot} setting, requiring no in-context demonstrations or question-SQL pairs. Evaluated on the BIRD benchmark with \texttt{Qwen2.5-72B-Instruct}, \textbf{SDE-SQL} achieves an \textbf{8.02\%} relative improvement in execution accuracy over the vanilla \texttt{Qwen2.5-72B-Instruct} baseline, establishing a new state-of-the-art among open-source methods without supervised fine-tuning (SFT) or model ensembling. Furthermore, when combined with SFT, \textbf{SDE-SQL} delivers an additional \textbf{0.52\%} performance gain.

\end{abstract}

\section{Introduction}
Text-to-SQL is a long-standing task in natural language processing that focuses on translating natural language questions into executable SQL queries. This capability not only empowers non-expert users to interact with structured databases seamlessly, but also mitigates hallucination issues in question-answering systems by grounding responses in factual, database-stored information.

Recent advances in large language models (LLMs) have led to significant improvements in the performance and accuracy of Text-to-SQL systems. LLM-based approaches have surpassed \textbf{90\%} execution accuracy on the original Spider dataset \citep{spider}, and have demonstrated promising results on more complex and diverse benchmarks such as BIRD\cite{bird}. Despite these advances, a noticeable gap remains between current model performance and human-level capabilities—particularly on the recently introduced Spider 2.0 benchmark \citep{spider2}, which poses more realistic and challenging scenarios for semantic parsing.
\begin{figure*}[t]
    \centering
    \includegraphics[width=1\linewidth]{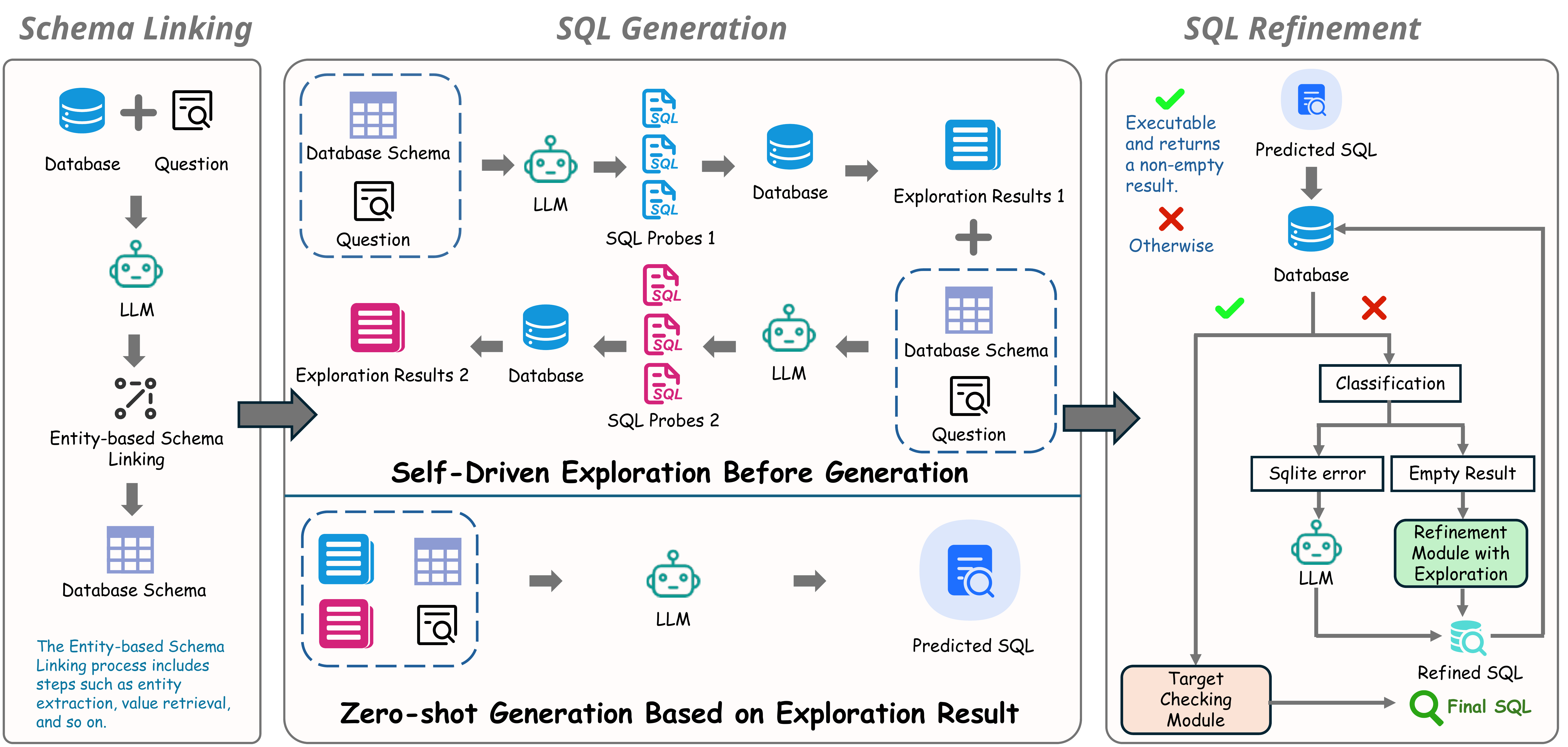}
    \caption{The Workflow of SDE-SQL, which consists of three parts: 1) Schema Linking: which
retrieves and selects useful database schema; 2)SQL Generation: performing zero-shot SQL generation based on two-phase self-driven exploration; 3)SQL Refinement: which refines the SQL with the execution results of the Sub-SQLs and SQL Probes. }
    \label{fig:workflow}
        \vspace{-3mm}
\end{figure*}
Contemporary large language model (LLM)-based approaches to Text-to-SQL typically comprise three core components: schema linking, SQL generation, and SQL refinement. In the schema linking stage, prior work has primarily focused on aligning natural language questions with relevant database schema elements, improving precision and contextual relevance. During SQL generation, various methods have been proposed to decompose complex questions and incorporate reasoning strategies. In the refinement stage, the categorization of SQL error types has become more systematic, enabling the development of targeted correction mechanisms.

Despite these advances, one crucial aspect of SQL remains largely underexplored: its inherent interactivity as a database interface that supports fast and informative execution. This underutilized property may partially account for the performance gap between LLM-based systems and human experts.

To address this, we propose \textbf{SDE-SQL}, a novel framework that incorporates \textbf{Self-Driven Exploration} into both the generation and refinement stages, as illustrated in Figure \ref{fig:workflow}. In addition to generating the final SQL query that directly answers the natural language question, the model autonomously generates and executes a sequence of auxiliary queries—termed \textbf{SQL Probes}—designed specifically to explore and extract informative signals from the database.

For schema linking, we leverage entity-based techniques including value retrieval and soft linking. During the generation phase, the model engages in a two-stage exploration process based on the question and schema, enabling it to iteratively refine its understanding of the database content and perform zero-shot reasoning grounded in the retrieved information.

Following generation, we incorporate a two-stage exploration process into the refinement phase. For SQL queries that return explicit execution errors, the model directly revises them based on the error feedback. For queries that execute successfully but return empty results, the first stage of exploration uses the execution results of decomposed sub-queries (\textit{Sub-SQLs}) to help the model diagnose the underlying issue. In the second stage, the model generates targeted \textbf{SQL Probes} to explore possible solutions, and selects the most promising one to produce the final refined query.

Empirically, \textbf{SDE-SQL} achieves an execution accuracy of \textbf{67.67\%} on the BIRD benchmark using \texttt{Qwen2.5-72B-Instruct} in a zero-shot setting. With supervised fine-tuning (SFT), the performance further improves to \textbf{68.19\%}.

\noindent \textbf{Our main contributions are as follows:}
\begin{itemize}
    \item We propose \textbf{SDE-SQL}, a novel framework that leverages \textbf{Self-Driven Exploration} to enhance the reasoning and interaction capabilities of LLMs in the Text-to-SQL task, significantly narrowing the gap with human experts.
    
    \item We introduce a unified exploration mechanism across both SQL generation and refinement stages, enabling LLMs to actively query the database, diagnose potential errors, and iteratively improve query quality.
    
    \item We conduct extensive experiments on the BIRD and Spider benchmarks, along with ablation studies, validating the effectiveness of Self-Driven Exploration.
    
    \item We build a small-scale dataset for supervised fine-tuning (SFT) on exploration and generation tasks, and show that targeted module-level fine-tuning further improves the performance of SDE-SQL.
\end{itemize}

\section{Related Work}
Transforming natural language questions into database queries is a classic task, the earliest works used inductive logic programming and human-designed templates to accomplish this task\citep{old_templates}. In recent years, the advancement of Text-to-SQL technologies can be broadly categorized into two stages, driven by progress in natural language processing. 
\subsection{Traditional Seq2Seq Model-Based Methods}
Previous work primarily focused on improving encoding or decoding methods, as the seq2seq model framework consists of two main components, the encoder and the decoder. IRNet employed a bidirectional LSTM to encode the question and a self-attention mechanism to encode the database schema, ultimately using an LSTM as a grammar-based decoder\citep{IRNet}. In order to effectively capture the relationship between the database schema and the question, RAT-SQL develops an encoder with a relation-aware self-attention mechanism\citep{RATSQL}. After that, \citet{SADGA} and \citet{LGESQL} utilized graph neural networks to encode the relationships between the schema and the query. Leveraging the exceptional capabilities of pre-trained language models (PLMs) across various NLP tasks, \citet{SQLova} was the first to incorporate BERT as its encoder. For improvements in the decoder, \citet{SQLNet} and \citet{RYANSQL} focused on sketch-
based decoding method. To reduce time consumption during inference, SDSQL presented the Schema Dependency Learning and removed execution-guided (EG) decoding strategy\citep{SDSQL}. 

\subsection{LLM-Based Methods}
With the advent of LLMs, the Text-to-SQL field has experienced a groundbreaking innovation, bringing about significant changes in the approach to the task.
\paragraph{Methods Based on Prompt Engineering }
\citet{eval_llm} evaluated the potential of LLMs in the Text-to-SQL task, demonstrating the remarkable capability of LLMs in this task. Building on in-context learning, DAIL-SQL \citep{DAILSQL} introduced a novel prompt engineering approach that improves the Text-to-SQL performance of LLMs through question representation, demonstration selection, and demonstration organization. Based on Chain-of-Thought(CoT) reasoning style\citep{CoT}, DIN-SQL\citep{DINSQL}, Divide-and-Prompt\citep{dvp}, CoE-SQL\citep{COESQL} and SQLfuse\citep{SQLfuse} designed CoT templates with reasoning steps in the prompt to elicit chain thinking. To enhance the ability of LLMs in handling complex problems, QDecomp\citep{QDecomp}, DIN-SQL\citep{DINSQL}, MAC-SQL\citep{MACSQL} and MAG-SQL\citep{MAGSQL} decomposed complex natural language questions and solve them step by step. Besides, MCS-SQL\citep{MCSSQL}, CHASE-SQL\citep{CHASESQL} and CHESS\citep{CHESS} enhanced performance by generating a large set of candidate SQL queries during the inference stage and selecting the most suitable ones.

\paragraph{Methods Based on Fine-tuning }
Although prompt engineering methods based on closed-source models, like GPT-4o\citep{gpt4o}, perform well in the Text-to-SQL task, they face issues such as high costs, inability to guarantee privacy, and limited flexibility. Therefore, fine-tuning open-source models for the Text-to-SQL task holds significant practical value and application potential.  DTS-SQL\citep{DTSSQL} and SQLfuse\citep{SQLfuse} explored fine-tuning LLMs for both schema linking and SQL generation. SQL-PaLM\citep{SQLPaLM}, Open-SQL\citep{OpenSQL}, XiYan-SQL\citep{XiYanSQL} and CodeS\citep{CodeS} fine-tuned open-source LLMs on carefully selected data, while CodeS specifically adopted an incremental pre-training approach using a specially curated SQL-centric corpus. In addition, there are some novel perspectives. DELLM\citet{DELLM} specifically fine-tuned a Data Expert Language Model that provides domain knowledge, while SQL-GEN\citet{SQLGEN} proposed a novel Mixture-of-Experts (MoE) architecture to handle multiple SQL dialects.

\section{Methodology}
\subsection{Entity-based Schema Linking}
In the Text-to-SQL task, schema linking refers to the process of identifying and selecting the relevant tables, columns, and values from the database based on the input natural language question. To improve the accuracy of linking, we use an entity-based linking approach, including \textbf{Value Retrieval} and \textbf{Soft Schema Linking}.
\subsubsection{Entity-based Value Retrieval}
Similar to the retrieval module in \citet{CHESS}, we first employ an LLM to extract entities from the natural language question through few-shot learning. And then the value retriever identifies similar
values in the database based on Locality Sensitive Hashing (LSH) and semantic similarity. 

\subsubsection{Entity-based Soft Schema Linking}
To improve the tolerance in the schema linking stage, we chose the soft schema linking method, like the approach in \citet{MAGSQL}. We employ a one-shot manner to prompt LLM to select the relevant columns based on each entity. For the selected columns, we provide as much detailed information as possible during the subsequent SQL generation, including the column name, type, column description, value examples, and value descriptions. For the unselected columns, we only retain the column name and type. This approach not only significantly reduces the input length, allowing the language model to focus on the most relevant database schema during generation, but also enhances tolerance by preventing the removal of useful columns that were not chosen. 

\subsection{Generation Based on Self-Driven Exploration}
\begin{figure}[h]
    \centering
    \includegraphics[width=1.0\linewidth]{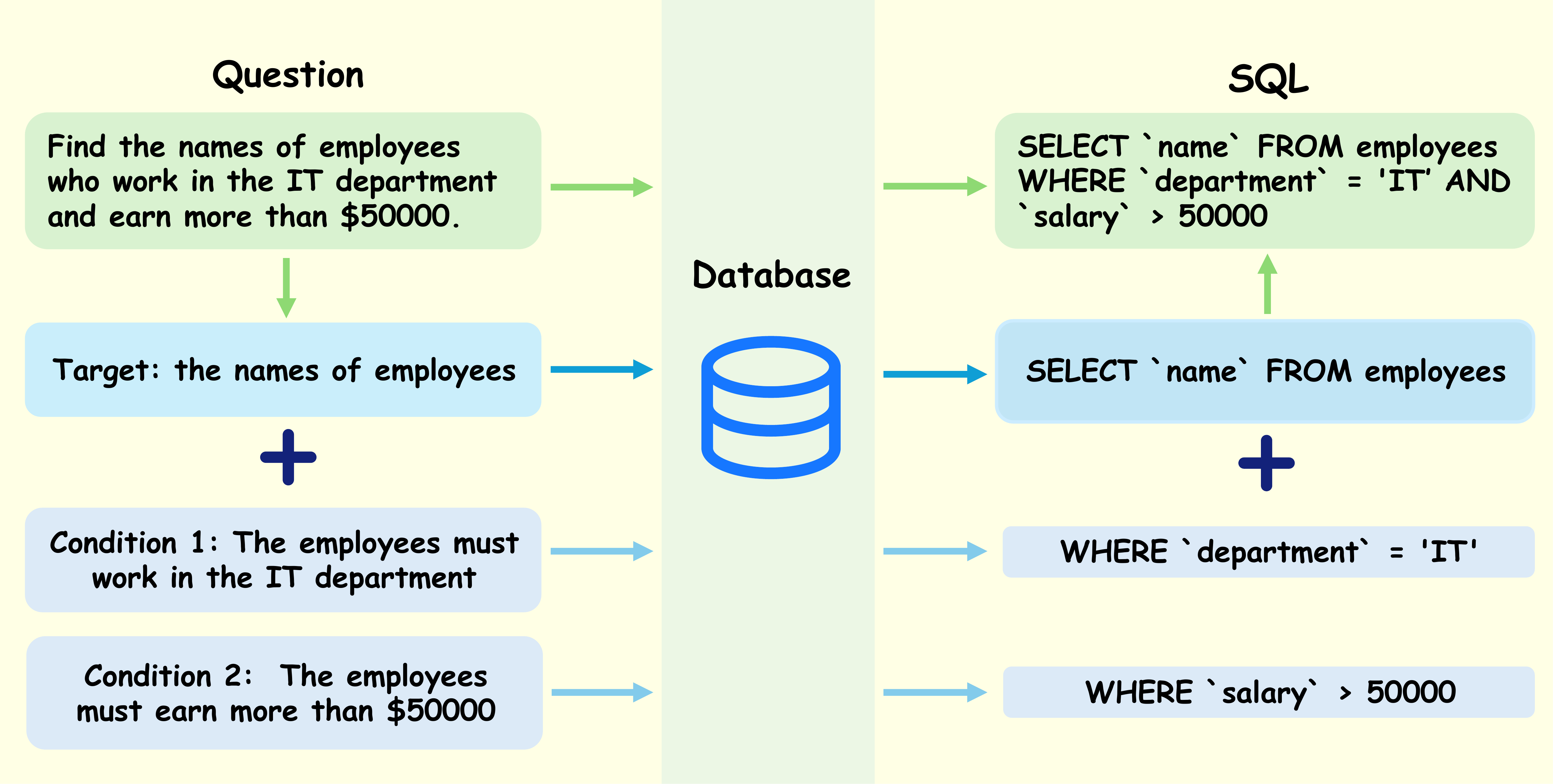}
    \caption{An example of Text-to-SQL.}
    \label{fig: Text-to-SQL}
\end{figure}

In previous Text-to-SQL research, SQL has often been viewed primarily as an intermediate result or final output, with its inherent functionality mostly overlooked. Therefore, we introduce the concept of \textbf{SQL Probes}. SQL Probes, literally meaning SQL queries that function as probes, are specifically designed for exploring the database based on current natural language question.
Formally, we define the task as a mapping from a natural language query \( Q \) and a database schema \( D \) to a corresponding SQL query \( S \). The natural language query \( Q \) is composed of two parts: the \textbf{target} and the \textbf{conditions} \citep{MAGSQL}. Typically, the target corresponds to the main \texttt{SELECT} clause in the SQL query \( S \), while the conditions correspond to the other clauses in \( S \), such as the \texttt{WHERE} clause. Figure \ref{fig: Text-to-SQL} is an example.

To obtain a specific SQL representation, entities must first be mapped to the corresponding columns and values in the database. Whether this step can be executed accurately depends on how well the language model understands the database. 

However, the information provided by the previously processed database schema is far from sufficient. Real-world databases are often highly complex and messy. Different tables may contain many columns with the same meaning (representing the same item), and the values in these columns might have different formats, with some values even existing only in specific tables. In the absence of sufficient information, LLM can only randomly identify combinations from these similar columns and values. This is also one of the key reasons behind the LLM's especially unstable performance in this task. During evaluation, it is frequently observed that the model can correctly predict some examples at times, while failing on the same examples at other times. In prior work, some methods have involved generating multiple SQL-candidates with the language model, followed by selecting the most appropriate one. Nevertheless, this approach fails to address the underlying problem. We propose that the most fundamental solution is to empower LLM with the ability to dynamically interact with the database. In SDE-SQL, LLM performs a two-stage self-driven exploration within the database before generation. 

\begin{figure*}[t]
    \centering
    \includegraphics[width=1\linewidth]{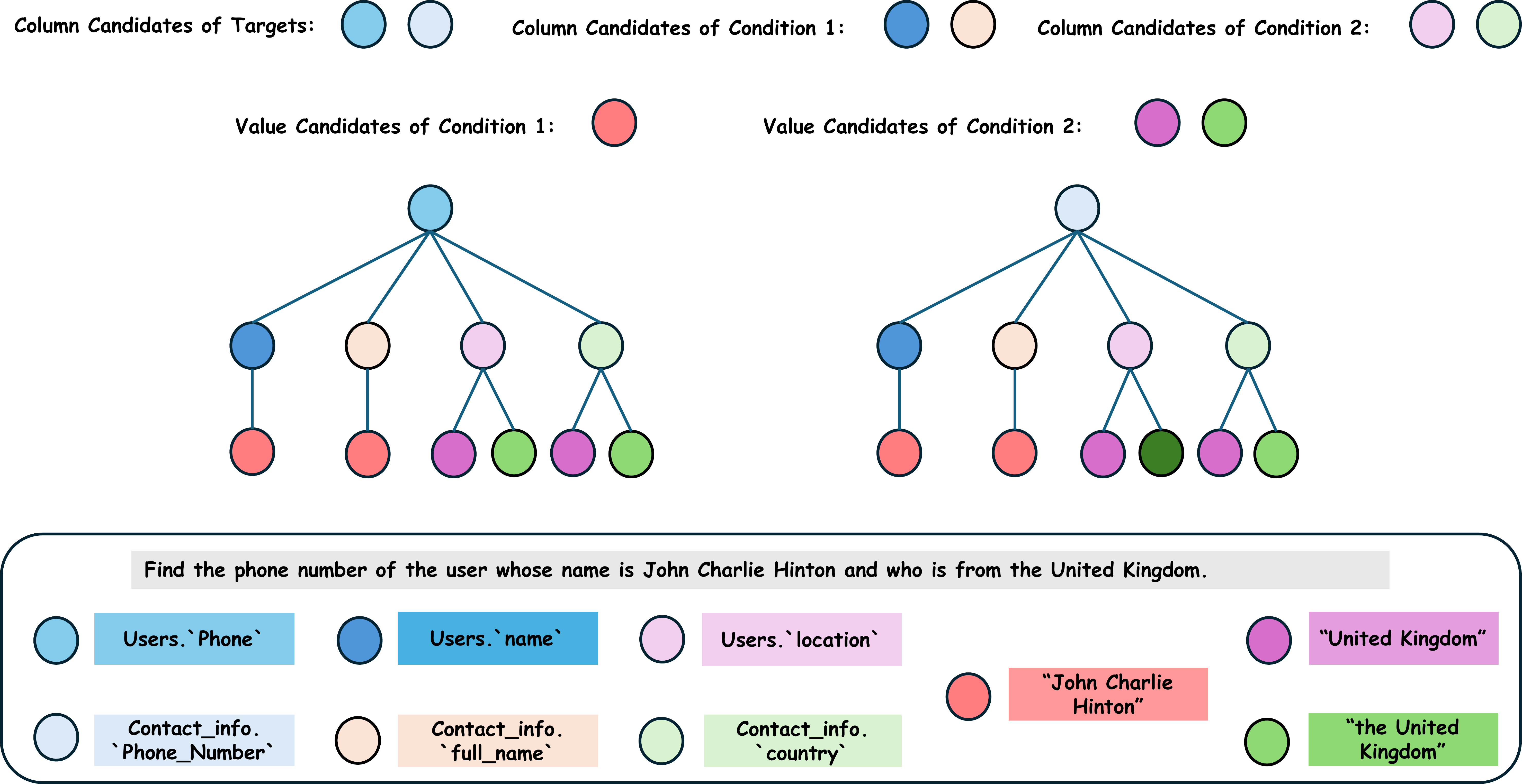}
    \caption{Condition SQL Probes Generation Process Illustrated Using a Tree Structure.}
    \label{fig: Probe Tree}
        \vspace{-3mm}
\end{figure*}

\subsubsection{Candidates Exploration}
The goal of this stage of exploration is to enable the large language model to query the database for information regarding both the Targets and a single Condition, and then select appropriate candidates for each target and condition. Since an entity in a natural language question is mapped to either a column or a value (or both a column and a value) in the database, LLM needs to determine the candidate columns and candidate values for each entity. Initially, the language model generates several \textbf{Base SQL Probes}, which enumerate candidate columns for the Targets. These SQLs focus solely on querying the Targets without any additional conditions. Following this, Condition SQL Probes are created, where each Probe typically extends a Base SQL Probe by adding a column candidate and maybe a value candidate corresponding to a specific condition. Assuming each set of candidates contains two options, the generation of Condition SQL Probes is illustrated in Figure \ref{fig: Probe Tree}, where each root-to-leaf path corresponds to a specific Condition SQL Probe. We refer to the condition description of each Condition SQL Probe as a \textbf{Condition Description Candidate}. For example, in Figure \ref{fig: Probe Tree}, one Condition Description Candidate is: 

\lstinline[language=SQL]|SELECT Phone FROM users WHERE name = 'John Charlie Hinton' AND location = 'United Kingdom';|

\subsubsection{Combinations Exploration}
Based on the results of the previous stage's exploration, the scope of candidates has been narrowed down. Now, it is necessary to combine all the conditions to find the most suitable candidate combination. For SQL queries that return no results, the corresponding candidate combination is definitely unsuitable.

\subsubsection{Zero-shot Generation with Exploration Results}
In our experiments, we found that existing methods do not fully leverage the large language model's potential for SQL generation. For example, strategies such as designing new decomposition approaches to allow the model to progressively solve complex problems, using various prompt techniques to generate multiple candidates for selection, or employing search strategies like Monte Carlo tree search(MCTS) to enhance the inference capability of language models, can lead to modest improvements in model performance. However, these gains are still significantly smaller than those achieved by providing the model with sufficient information. 

Therefore, in SDE-SQL, the LLM generator generates SQL based on the database schema and the results from the previous two exploration stages, without relying on any question-SQL pairs as few-shot examples or using any question decomposition strategies. To improve the accuracy and robustness of SQL generation, we adopt a self-consistency strategy that selects the most consistent answer by comparing the execution results of multiple generated SQL queries.

\begin{figure*}[t]
    \centering
    \includegraphics[width=0.8\linewidth]{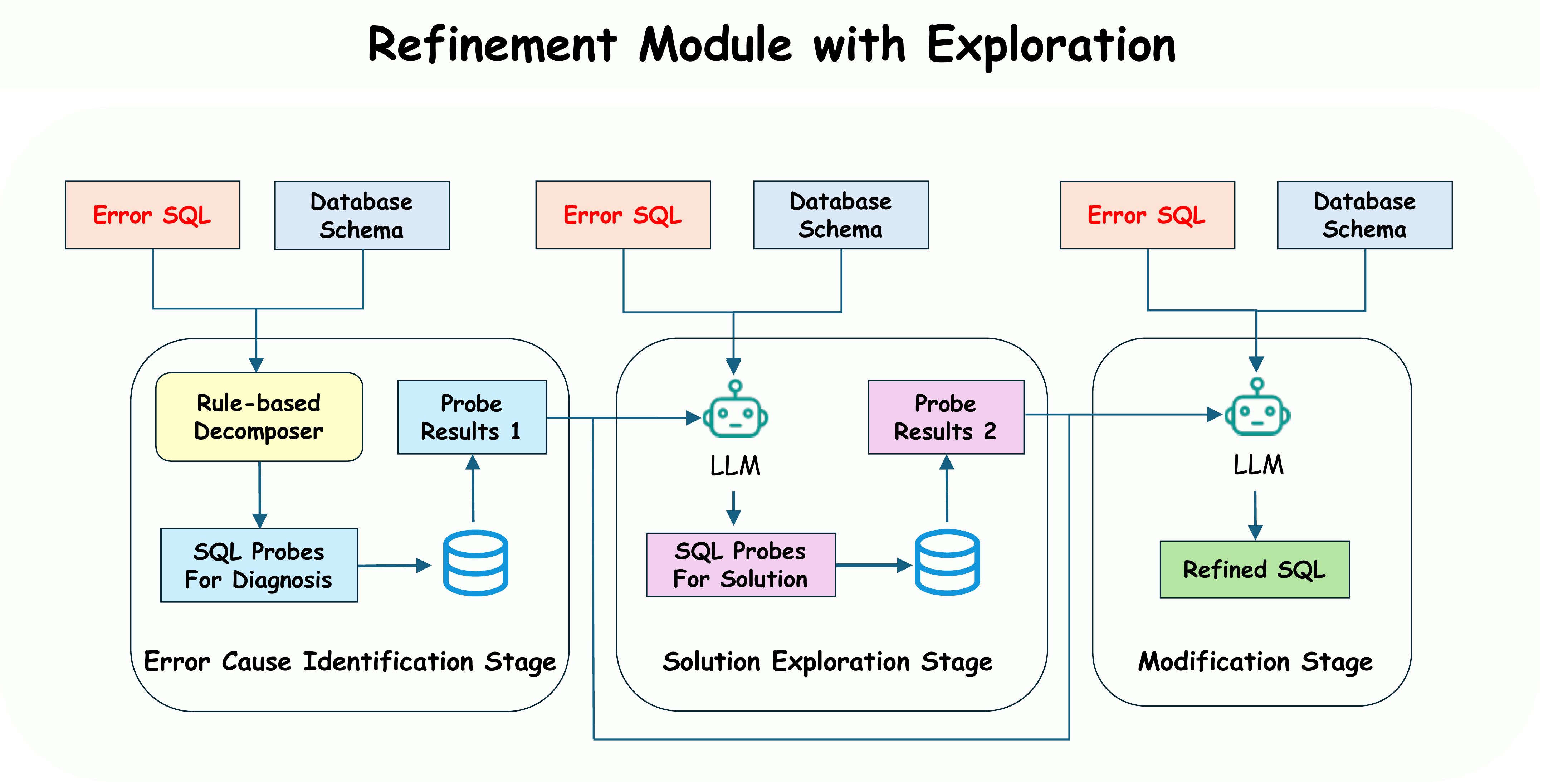}
    \caption{An illustration of the proposed refinement process with exploration in SDE-SQL. }
    \label{fig:refinement}
    \vspace{-3mm}
\end{figure*}

\subsection{Refinement Based on Self-Driven Exploration}
In the past, existing techniques based on In-Context Learning have introduced detection and repair solutions for Text-to-SQL errors, with each solution differing in its approach to error identification algorithms and the supplementary data provided to assist LLM in comprehending and rectifying these errors.\citep{maplerepair} 

For Syntax errors and Schema errors, the error feedback after execution already contains sufficient information, allowing LLMs to effectively complete the correction of SQL. However, for some other more complex errors, they typically result in empty query results without any error messages. Even when humans attempt to correct these errors, they cannot do so in one go; instead, they need to write some SQL statements for debugging and diagnose the problem based on the execution results of these queries. The current approach involves continuously regenerating until the repairs is successful or the attempt limit is reached. Throughout this repair process, LLM does not receive any useful information, and its reasoning abilities are not fully utilized. In other words, the reason for the error is never identified.

Therefore, in \textbf{SDE-SQL}, we introduce a comprehensive \textbf{Self-Driven Exploration} phase prior to SQL revision. For queries that yield empty results, the refinement process is divided into three distinct stages: the \textbf{Error Cause Identification Stage}, the \textbf{Solution Exploration Stage}, and the \textbf{Modification Stage}, as illustrated in Figure~\ref{fig:refinement}.

\subsubsection{Error Cause Identification Stage}
In a complex SQL statement, multiple tables may be involved and multiple conditions may be applied simultaneously, making it difficult to pinpoint the issue by directly analyzing the entire SQL. Therefore, we need to conduct a fine-grained diagnosis. To generate a series of Sub-SQLs as SQL Probes for diagnostic purposes, we developed a decomposer based on \textbf{SQLGlot}. The decomposer first converts complex SQL queries into \textbf{Abstract Syntax Trees (ASTs)} and then identifies indivisible condition units by analyzing node types and their relationships. These identified subtrees within the AST serve as the foundation for generating semantically valid Sub-SQLs, and the execution results of these Sub-SQLs will be provided to LLM to assist it in accurately diagnosing and pinpointing issues in the original query. An example of the decomposition result is shown in Figure \ref{fig:error_search}.

\subsubsection{Solution Exploration Stage}
We have summarized five possible reasons that may lead to an empty query result. At this stage, LLM need to analyze the probe results in the previous stage to derive hypotheses about possible error causes, and then generate a series of SQL probes to assist in exploring potential solutions to these errors. 
\paragraph{Conditions conflict or condition duplication}
This error refers to situations where data can be found when executed under a single condition, but when multiple conditions are combined, no data that meets the requirements can be found (resulting in an empty query result). There are two possible reasons for this error: conflicting combinations of multiple different conditions or redundant descriptions of a single condition using different columns. (\romannumeral1) Conditions conflict typically arises when an entity in a condition corresponds to multiple possible candidate columns, and only a specific candidate column combined with other conditions can yield the corresponding data item. An example is shown in Figure \ref{fig:condition_conflict}. (\romannumeral2) Condition duplication occurs when an entity in a condition maps to multiple candidate columns, causing the SQL generated by the large language model to inadvertently employ these various candidate columns in describing the same condition, ultimately resulting in the failure to retrieve data that fulfills the intended condition.

\paragraph{Unnecessary Table Joins}
The SQL may include unnecessary table joins, resulting in no records satisfying the conditions in the final intersection.
\paragraph{Mismatch between column and value}
This error arises when either the value format does not match the selected (but correct) column, or when a similar-looking column is chosen that does not contain the intended value.
\paragraph{Sub-query Scope Inconsistency}
Sometimes, the scope of the sub-query may be inconsistent with that of the main query, especially when using the \texttt{MIN/MAX} functions in the sub-query, which often leads to an empty query result. Figure \ref{fig:subquery_scope} shows an example that the row of data retrieved in the sub-query does not exist in the result after the JOIN of these two tables.

\subsubsection{Target Checking After Refinement}
For an SQL query, the most important part is actually the target of the query, which refers to the columns being selected in the \texttt{SELECT} clause. If the query target in SQL does not align with the original query target in the natural language question, then the transformation is undoubtedly a failure. However, when LLMs generate SQL, they sometimes include columns that are not part of the query target in the \texttt{SELECT} clause, such as columns used in the conditions. Therefore, after refining the SQL based on the execution results, it is necessary to check whether the query target in the SQL matches the query target in the natural language question. To avoid introducing new errors at this stage, we allow the large language model to only determine if unnecessary target columns are selected in the SQL. If such columns are found, they will be removed without affecting the execution. The procedure is illustrated in Figure \ref{fig:target_check}.

\subsection{Supervised Fine-Tuning (SFT)}
To further enhance the model’s ability to autonomously explore the database and utilize exploration results to generate more accurate SQL, we also perform supervised fine-tuning (SFT) on the model. The training data is sampled from the training set of BIRD. We employed a prompt-based pipeline to roll out data, and the examples that eventually produced correct SQL were regarded as valid data for fine-tuning the model. 

Among the 9,428 data points, 5,231 valid samples were obtained through sampling. From the reasoning trajectory of each example, we extract two components: ({\romannumeral1}) the exploration phase, where SQL probes are generated; and ({\romannumeral2}) the prediction phase, where the final SQL query is generated based on the exploration results. 

\section{Experiments}
In this section, we first introduce the experimental setup, and then report and analyze the results.
\subsection{Experimental Setup}
\subsubsection{Dataset and Metrics}
Spider \cite{spider} is a widely adopted benchmark dataset for the Text-to-SQL task. It is large-scale, cross-domain, and complex, containing 10,181 natural language questions and 5,693 corresponding SQL queries across 200 different databases. As a challenging benchmark of Text-to-SQL task, the recently proposed BIRD dataset \cite{bird} includes 95 large-scale real databases with dirty values, featuring 12,751 unique question-SQL pairs. The databases within the BIRD dataset, similar to those in real-world scenarios, exhibit inherent ambiguities. Accordingly, detailed descriptions are provided for each column, along with external knowledge. In this work, we choose Execution Accuracy (EX) as the metric, since it reflects the accuracy of the results returned by the executed SQL queries. This metric considers various SQL formulations for the same question, providing a more precise and fair evaluation of the outcomes.
\begin{table*}[t]
  \renewcommand{\arraystretch}{1.2}
  \centering
  \small
  \begin{tabular}{lcccc}
    \hline
    \textbf{Method}           & \textbf{Simple} & \textbf{Moderate}  & \textbf{Challenging} &\textbf{All} \\
    \hline
    \textbf{SDE-SQL + Qwen2.5-72B-Instruct}       & \textbf{74.92}     &\textbf{57.76}    &\textbf{53.10}  & \textbf{67.67}\\
    \quad w/o Soft Schema Linker &73.51                &58.84                &50.34               &66.88\textsubscript{\scriptsize\color{red}$\downarrow$0.79}  \\
    \quad w/o Exploration Before Generation &72.97 &56.46 &48.97 &65.71\textsubscript{\scriptsize\color{red}$\downarrow$1.96} \\
    \quad w/o Refinement Module &72.97                &55.60                &48.97                &65.45\textsubscript{\scriptsize\color{red}$\downarrow$2.22} \\
    \quad \quad w/o Exploration in Refinement &72.86 &56.68 &51.72 & 65.97\textsubscript{\scriptsize\color{red}$\downarrow$1.70}\\
    \quad \quad w/o Target Checking &73.19 &57.76 &49.66 &66.30 \textsubscript{\scriptsize\color{red}$\downarrow$1.37}\\
    \quad w/o Exploration in Generation \& Refinement &72.11                &54.31                &48.28                &64.47\textsubscript{\scriptsize\color{red}$\downarrow$3.20} \\
\hline
     SDE-SQL + Fine-tuned Explorer       &74.16                &59.26                &55.17                &67.86\textsubscript{\scriptsize\color{darkgreen}$\uparrow$0.19}\\
         
     SDE-SQL + Fine-tuned Generator       &74.49                &58.19                &55.86                &67.80\textsubscript{\scriptsize\color{darkgreen}$\uparrow$0.13}\\
     \textbf{SDE-SQL + Fine-tuned Explorer \& Generator}       & \textbf{74.70}     &\textbf{58.84}    &\textbf{56.55}  & \textbf{68.19}\textsubscript{\scriptsize\color{darkgreen}$\uparrow$0.52}\\
     \hline
  \end{tabular}
  \caption{Execution accuracy of SDE-SQL on BIRD dev set in the ablation study.}
  \label{agents_ablation}
\end{table*}
\subsubsection{SFT Settings}
For both exploration task and generation task, we conducted 24-hour training on 8 NVIDIA A800 GPUs with Qwen2.5-72B-Instruct. The detailed training hyperparameters are provided in Table \ref{tab:train}.
\subsubsection{Baselines}
To enable a comprehensive comparison, we selected representative methods based on \textbf{closed-source} models and representative methods based on \textbf{open-source} models \textbf{without model ensemble} as baselines. 

\subsection{Main Results}
\subsubsection{BIRD Results}
When evaluated on the BIRD dev dataset, \textbf{SDE-SQL} based on Qwen2.5-72B-Instruct outperforms \textbf{most GPT-4-based methods} and the majority of open-source models, achieving an execution accuracy of \textbf{68.19\%} after fine-tuning, as shown in Table~\ref{BIRD_Results}. Even in the training-free setting, it achieves a strong performance of \textbf{67.67\%}, further highlighting the effectiveness of our approach.
\begin{table}[h]
  \renewcommand{\arraystretch}{1.2}
  \centering
  \small
  \begin{tabular}{lc}
    \hline
    \textbf{Method}           & \textbf{dev(EX)} \\
    \hline
      AskData + GPT-4o & \textbf{75.36} \\
      CHASE-SQL + Gemini & 74.46 \\
    XiYan-SQL & 73.34 \\
       OpenSearch-SQL, v2 + GPT-4o & 69.30 \\
      CHESS & 68.31 \\
      Distillery + GPT-4o & 67.21 \\
      MCS-SQL & 63.36 \\
    MAC-SQL + GPT-4    & 59.39  \\
    DAIL-SQL + GPT-4 & 54.76 \\
    DIN-SQL + GPT-4 & 50.72 \\
    GPT-4 & 46.35 \\
    \hline 
    DTS-SQL + DeepSeek-7B & 55.80 \\
     SFT CodeS-15B & 58.47 \\
     SQL-o1 + Llama3-8B & 63.4 \\
    OneSQL-v0.1-Qwen-32B & 64.60 \\
    XiYanSQL-QwenCoder-32B & 67.01\\
            Qwen2.5-72B-Instruct & 60.17 \\
    \hline
    \textbf{SDE-SQL + Qwen2.5-72B-Instruct} & \textbf{67.67} \\
    \textbf{SDE-SQL (SFT)}    & \textbf{68.19}  \\
    
    \hline
  \end{tabular}
  \caption{The experimental results of competing model on the   BIRD dataset. }
  \label{BIRD_Results}
\end{table}
\subsubsection{Spider Results}
As shown in Table \ref{Spider_results}, SDE-SQL fine-tuned solely on the BIRD training set achieves competitive results on Spider benchmark, surpassing GPT-4-based methods and most open-source models, which underscores its strong generalization ability. Nevertheless, the performance gain is relatively modest, as a large portion of SQL queries in the Spider dataset produce empty execution results, thereby limiting the effectiveness of feedback from database exploration.
\begin{table}[h]
  \renewcommand{\arraystretch}{1.2}
  \centering
  \small 
  \resizebox{\linewidth}{!}{
  \begin{tabular}{lcc}
    \hline
    \textbf{Method}           & \textbf{dev(EX)} & \textbf{test(EX)}\\
    \hline
    
    \textbf{SDE-SQL (SFT)}    & \textbf{87.5}  & \textbf{88.5}\\
    \textbf{SDE-SQL + Qwen2.5-72B-Instruct}    & \textbf{87.3}  & \textbf{88.3}\\
    MAC-SQL + GPT-4    & {86.8}  & 82.8\\
    SENSE-13B &84.1 &83.5 \\
    SQL-o1 + Llama3-8B &87.4 &85.4 \\
    DAIL-SQL + GPT-4  & 84.4  & 86.6\\
    ROUTE + Qwen2.5-14B & 87.3 & 87.1 \\
    DIN-SQL + GPT-4 & 82.8 & 85.3\\
    GPT-4 (zero-shot) & 73.4 & - \\
    Qwen2.5-72B-Instruct & 73.9 & 84.0 \\
    \hline
  \end{tabular}
  }
  \caption{ The experimental results of competing model on the   Spider dataset.}
  \label{Spider_results}
\end{table}
\subsection{Ablation Study}
For each module in SDE-SQL, we conduct ablation studies on the development set of BIRD benchmark, which is shown in Table \ref{agents_ablation}. In addition, we evaluate the effect of incorporating the fine-tuned explorer and generator into the pipeline. The results demonstrate that each component plays an important role, with the introduction of the two exploration phases leading to particularly significant performance improvements. Besides, modules fine-tuned on their respective sub-tasks can further enhance the overall performance of the workflow.

\section{Conclusion}
In this work, we propose \textbf{SDE-SQL}, a novel Text-to-SQL framework that integrates \emph{Self-Driven Exploration} into both the SQL generation and refinement stages. By enabling LLMs to proactively interact with databases through SQL probes, SDE-SQL bridges the gap between static query generation and dynamic, execution-based reasoning. This exploration mechanism allows LLMs to uncover latent schema semantics and execution patterns, significantly improving their ability to produce executable and semantically accurate SQL queries. Extensive experiments on the BIRD and Spider datasets demonstrate the effectiveness of SDE-SQL, with the model achieving an execution accuracy of \textbf{68.19\%} on BIRD after supervised fine-tuning. Ablation studies confirm the contributions of key components---especially the exploration pipeline and the fine-tuning strategies. As future work, we plan to explore tighter integration of exploration signals into model training to further strengthen the model’s reasoning capabilities.

\section{Limitation}
Although self-driven exploration significantly enhances the potential of large language models in Text-to-SQL tasks, our current approach has several limitations. In SDE-SQL, database exploration is entirely prompt-driven—meaning that the effectiveness of the exploration process heavily depends on the design and quality of manually crafted prompts. Poorly constructed prompts may lead the model to generate uninformative or redundant SQL probes, thereby limiting its ability to acquire meaningful schema knowledge or execution insights. Moreover, relying solely on prompt engineering can restrict the model’s capacity for deeper reasoning, as it lacks mechanisms for adaptive learning based on feedback from the environment.

Another limitation is the model’s inability to autonomously refine its exploration strategy over time. Since each SQL probe is generated statically from prompts, the model cannot dynamically adjust its behavior based on prior successes or failures during the exploration process. This constraint reduces the overall flexibility and learning efficiency of the system.

To address these issues, future work will focus on making database exploration more intrinsic to the model itself. One promising direction is to incorporate reinforcement learning or other feedback-driven learning paradigms, allowing the model to iteratively refine its probing strategies based on execution outcomes. By enabling the model to learn from its own interactions with the database, we hope to develop a more robust, adaptive framework capable of deeper, context-aware reasoning in complex database environments.



\bibliography{custom}

\begin{thebibliography}{34}
\providecommand{\natexlab}[1]{#1}

\bibitem[{Cai et~al.(2022)Cai, Yuan, Xu, and Hao}]{SADGA}
Ruichu Cai, Jinjie Yuan, Boyan Xu, and Zhifeng Hao. 2022.
\newblock \href {https://arxiv.org/abs/2111.00653} {Sadga: Structure-aware dual graph aggregation network for text-to-sql}.
\newblock \emph{Preprint}, arXiv:2111.00653.

\bibitem[{Cao et~al.(2021)Cao, Chen, Chen, Zhao, Zhu, and Yu}]{LGESQL}
Ruisheng Cao, Lu~Chen, Zhi Chen, Yanbin Zhao, Su~Zhu, and Kai Yu. 2021.
\newblock \href {https://arxiv.org/abs/2106.01093} {Lgesql: Line graph enhanced text-to-sql model with mixed local and non-local relations}.
\newblock \emph{Preprint}, arXiv:2106.01093.

\bibitem[{Chen et~al.(2024)Chen, Wang, Qiu, Qin, and Yang}]{OpenSQL}
Xiaojun Chen, Tianle Wang, Tianhao Qiu, Jianbin Qin, and Min Yang. 2024.
\newblock \href {https://arxiv.org/abs/2405.06674} {Open-sql framework: Enhancing text-to-sql on open-source large language models}.
\newblock \emph{Preprint}, arXiv:2405.06674.

\bibitem[{Choi et~al.(2020)Choi, Shin, Kim, and Shin}]{RYANSQL}
DongHyun Choi, Myeong~Cheol Shin, EungGyun Kim, and Dong~Ryeol Shin. 2020.
\newblock \href {https://arxiv.org/abs/2004.03125} {Ryansql: Recursively applying sketch-based slot fillings for complex text-to-sql in cross-domain databases}.
\newblock \emph{Preprint}, arXiv:2004.03125.

\bibitem[{Gao et~al.(2023)Gao, Wang, Li, Sun, Qian, Ding, and Zhou}]{DAILSQL}
Dawei Gao, Haibin Wang, Yaliang Li, Xiuyu Sun, Yichen Qian, Bolin Ding, and Jingren Zhou. 2023.
\newblock \href {https://arxiv.org/abs/2308.15363} {Text-to-sql empowered by large language models: A benchmark evaluation}.
\newblock \emph{Preprint}, arXiv:2308.15363.

\bibitem[{Gao et~al.(2025)Gao, Liu, Li, Shi, Zhu, Wang, Li, Li, Hong, Luo, Gao, Mou, and Li}]{XiYanSQL}
Yingqi Gao, Yifu Liu, Xiaoxia Li, Xiaorong Shi, Yin Zhu, Yiming Wang, Shiqi Li, Wei Li, Yuntao Hong, Zhiling Luo, Jinyang Gao, Liyu Mou, and Yu~Li. 2025.
\newblock \href {https://arxiv.org/abs/2411.08599} {A preview of xiyan-sql: A multi-generator ensemble framework for text-to-sql}.
\newblock \emph{Preprint}, arXiv:2411.08599.

\bibitem[{Guo et~al.(2019)Guo, Zhan, Gao, Xiao, Lou, Liu, and Zhang}]{IRNet}
Jiaqi Guo, Zecheng Zhan, Yan Gao, Yan Xiao, Jian-Guang Lou, Ting Liu, and Dongmei Zhang. 2019.
\newblock \href {https://arxiv.org/abs/1905.08205} {Towards complex text-to-sql in cross-domain database with intermediate representation}.
\newblock \emph{Preprint}, arXiv:1905.08205.

\bibitem[{Hong et~al.(2024)Hong, Yuan, Chen, Zhang, Huang, and Huang}]{DELLM}
Zijin Hong, Zheng Yuan, Hao Chen, Qinggang Zhang, Feiran Huang, and Xiao Huang. 2024.
\newblock \href {https://arxiv.org/abs/2402.11517} {Knowledge-to-sql: Enhancing sql generation with data expert llm}.
\newblock \emph{Preprint}, arXiv:2402.11517.

\bibitem[{Hui et~al.(2021)Hui, Shi, Geng, Li, Li, Sun, and Zhu}]{SDSQL}
Binyuan Hui, Xiang Shi, Ruiying Geng, Binhua Li, Yongbin Li, Jian Sun, and Xiaodan Zhu. 2021.
\newblock \href {https://arxiv.org/abs/2103.04399} {Improving text-to-sql with schema dependency learning}.
\newblock \emph{Preprint}, arXiv:2103.04399.

\bibitem[{Hwang et~al.(2019)Hwang, Yim, Park, and Seo}]{SQLova}
Wonseok Hwang, Jinyeong Yim, Seunghyun Park, and Minjoon Seo. 2019.
\newblock \href {https://arxiv.org/abs/1902.01069} {A comprehensive exploration on wikisql with table-aware word contextualization}.
\newblock \emph{Preprint}, arXiv:1902.01069.

\bibitem[{Lee et~al.(2024)Lee, Park, Kim, and Park}]{MCSSQL}
Dongjun Lee, Choongwon Park, Jaehyuk Kim, and Heesoo Park. 2024.
\newblock \href {https://arxiv.org/abs/2405.07467} {Mcs-sql: Leveraging multiple prompts and multiple-choice selection for text-to-sql generation}.
\newblock \emph{Preprint}, arXiv:2405.07467.

\bibitem[{Lei et~al.(2025)Lei, Chen, Ye, Cao, Shin, Su, Suo, Gao, Hu, Yin, Zhong, Xiong, Sun, Liu, Wang, and Yu}]{spider2}
Fangyu Lei, Jixuan Chen, Yuxiao Ye, Ruisheng Cao, Dongchan Shin, Hongjin Su, Zhaoqing Suo, Hongcheng Gao, Wenjing Hu, Pengcheng Yin, Victor Zhong, Caiming Xiong, Ruoxi Sun, Qian Liu, Sida Wang, and Tao Yu. 2025.
\newblock \href {https://arxiv.org/abs/2411.07763} {Spider 2.0: Evaluating language models on real-world enterprise text-to-sql workflows}.
\newblock \emph{Preprint}, arXiv:2411.07763.

\bibitem[{Li et~al.(2024)Li, Zhang, Liu, Fan, Zhang, Zhu, Wei, Pan, Li, and Chen}]{CodeS}
Haoyang Li, Jing Zhang, Hanbing Liu, Ju~Fan, Xiaokang Zhang, Jun Zhu, Renjie Wei, Hongyan Pan, Cuiping Li, and Hong Chen. 2024.
\newblock \href {https://arxiv.org/abs/2402.16347} {Codes: Towards building open-source language models for text-to-sql}.
\newblock \emph{Preprint}, arXiv:2402.16347.

\bibitem[{Li et~al.(2023)Li, Hui, Qu, Yang, Li, Li, Wang, Qin, Cao, Geng, Huo, Zhou, Ma, Li, Chang, Huang, Cheng, and Li}]{bird}
Jinyang Li, Binyuan Hui, Ge~Qu, Jiaxi Yang, Binhua Li, Bowen Li, Bailin Wang, Bowen Qin, Rongyu Cao, Ruiying Geng, Nan Huo, Xuanhe Zhou, Chenhao Ma, Guoliang Li, Kevin C.~C. Chang, Fei Huang, Reynold Cheng, and Yongbin Li. 2023.
\newblock \href {https://arxiv.org/abs/2305.03111} {Can llm already serve as a database interface? a big bench for large-scale database grounded text-to-sqls}.
\newblock \emph{Preprint}, arXiv:2305.03111.

\bibitem[{Liu and Tan(2023)}]{dvp}
Xiping Liu and Zhao Tan. 2023.
\newblock \href {https://arxiv.org/abs/2304.11556} {Divide and prompt: Chain of thought prompting for text-to-sql}.
\newblock \emph{Preprint}, arXiv:2304.11556.

\bibitem[{OpenAI et~al.(2024)OpenAI, :, Hurst, Lerer, Goucher, Perelman, Ramesh, Clark, Ostrow, Welihinda, Hayes, Radford, Madry, Baker-Whitcomb, Beutel, Borzunov, Carney, Chow, Kirillov, Nichol, Paino, Renzin, Passos, Kirillov, Christakis, Conneau, Kamali, Jabri, Moyer, Tam, Crookes, Tootoochian, Tootoonchian, Kumar, Vallone, Karpathy, Braunstein, Cann, Codispoti, Galu, Kondrich, Tulloch, Mishchenko, Baek, Jiang, Pelisse, Woodford, Gosalia, Dhar, Pantuliano, Nayak, Oliver, Zoph, Ghorbani, Leimberger, Rossen, Sokolowsky, Wang, Zweig, Hoover, Samic, McGrew, Spero, Giertler, Cheng, Lightcap, Walkin, Quinn, Guarraci, Hsu, Kellogg, Eastman, Lugaresi, Wainwright, Bassin, Hudson, Chu, Nelson, Li, Shern, Conger, Barette, Voss, Ding, Lu, Zhang, Beaumont, Hallacy, Koch, Gibson, Kim, Choi, McLeavey, Hesse, Fischer, Winter, Czarnecki, Jarvis, Wei, Koumouzelis, Sherburn, Kappler, Levin, Levy, Carr, Farhi, Mely, Robinson, Sasaki, Jin, Valladares, Tsipras, Li, Nguyen, Findlay, Oiwoh, Wong, Asdar, Proehl, Yang, Antonow,
  Kramer, Peterson, Sigler, Wallace, Brevdo, Mays, Khorasani, Such, Raso, Zhang, von Lohmann, Sulit, Goh, Oden, Salmon, Starace, Brockman, Salman, Bao, Hu, Wong, Wang, Schmidt, Whitney, Jun, Kirchner, de~Oliveira~Pinto, Ren, Chang, Chung, Kivlichan, O'Connell, O'Connell, Osband, Silber, Sohl, Okuyucu, Lan, Kostrikov, Sutskever, Kanitscheider, Gulrajani, Coxon, Menick, Pachocki, Aung, Betker, Crooks, Lennon, Kiros, Leike, Park, Kwon, Phang, Teplitz, Wei, Wolfe, Chen, Harris, Varavva, Lee, Shieh, Lin, Yu, Weng, Tang, Yu, Jang, Candela, Beutler, Landers, Parish, Heidecke, Schulman, Lachman, McKay, Uesato, Ward, Kim, Huizinga, Sitkin, Kraaijeveld, Gross, Kaplan, Snyder, Achiam, Jiao, Lee, Zhuang, Harriman, Fricke, Hayashi, Singhal, Shi, Karthik, Wood, Rimbach, Hsu, Nguyen, Gu-Lemberg, Button, Liu, Howe, Muthukumar, Luther, Ahmad, Kai, Itow, Workman, Pathak, Chen, Jing, Guy, Fedus, Zhou, Mamitsuka, Weng, McCallum, Held, Ouyang, Feuvrier, Zhang, Kondraciuk, Kaiser, Hewitt, Metz, Doshi, Aflak, Simens, Boyd,
  Thompson, Dukhan, Chen, Gray, Hudnall, Zhang, Aljubeh, Litwin, Zeng, Johnson, Shetty, Gupta, Shah, Yatbaz, Yang, Zhong, Glaese, Chen, Janner, Lampe, Petrov, Wu, Wang, Fradin, Pokrass, Castro, de~Castro, Pavlov, Brundage, Wang, Khan, Murati, Bavarian, Lin, Yesildal, Soto, Gimelshein, Cone, Staudacher, Summers, LaFontaine, Chowdhury, Ryder, Stathas, Turley, Tezak, Felix, Kudige, Keskar, Deutsch, Bundick, Puckett, Nachum, Okelola, Boiko, Murk, Jaffe, Watkins, Godement, Campbell-Moore, Chao, McMillan, Belov, Su, Bak, Bakkum, Deng, Dolan, Hoeschele, Welinder, Tillet, Pronin, Tillet, Dhariwal, Yuan, Dias, Lim, Arora, Troll, Lin, Lopes, Puri, Miyara, Leike, Gaubert, Zamani, Wang, Donnelly, Honsby, Smith, Sahai, Ramchandani, Huet, Carmichael, Zellers, Chen, Chen, Nigmatullin, Cheu, Jain, Altman, Schoenholz, Toizer, Miserendino, Agarwal, Culver, Ethersmith, Gray, Grove, Metzger, Hermani, Jain, Zhao, Wu, Jomoto, Wu, Shuaiqi, Xia, Phene, Papay, Narayanan, Coffey, Lee, Hall, Balaji, Broda, Stramer, Xu, Gogineni,
  Christianson, Sanders, Patwardhan, Cunninghman, Degry, Dimson, Raoux, Shadwell, Zheng, Underwood, Markov, Sherbakov, Rubin, Stasi, Kaftan, Heywood, Peterson, Walters, Eloundou, Qi, Moeller, Monaco, Kuo, Fomenko, Chang, Zheng, Zhou, Manassra, Sheu, Zaremba, Patil, Qian, Kim, Cheng, Zhang, He, Zhang, Jin, Dai, and Malkov}]{gpt4o}
OpenAI, :, Aaron Hurst, Adam Lerer, Adam~P. Goucher, Adam Perelman, Aditya Ramesh, Aidan Clark, AJ~Ostrow, Akila Welihinda, Alan Hayes, Alec Radford, Aleksander Madry, Alex Baker-Whitcomb, Alex Beutel, Alex Borzunov, Alex Carney, Alex Chow, Alex Kirillov, Alex Nichol, Alex Paino, Alex Renzin, Alex~Tachard Passos, Alexander Kirillov, Alexi Christakis, Alexis Conneau, Ali Kamali, Allan Jabri, Allison Moyer, Allison Tam, Amadou Crookes, Amin Tootoochian, Amin Tootoonchian, Ananya Kumar, Andrea Vallone, Andrej Karpathy, Andrew Braunstein, Andrew Cann, Andrew Codispoti, Andrew Galu, Andrew Kondrich, Andrew Tulloch, Andrey Mishchenko, Angela Baek, Angela Jiang, Antoine Pelisse, Antonia Woodford, Anuj Gosalia, Arka Dhar, Ashley Pantuliano, Avi Nayak, Avital Oliver, Barret Zoph, Behrooz Ghorbani, Ben Leimberger, Ben Rossen, Ben Sokolowsky, Ben Wang, Benjamin Zweig, Beth Hoover, Blake Samic, Bob McGrew, Bobby Spero, Bogo Giertler, Bowen Cheng, Brad Lightcap, Brandon Walkin, Brendan Quinn, Brian Guarraci, Brian Hsu,
  Bright Kellogg, Brydon Eastman, Camillo Lugaresi, Carroll Wainwright, Cary Bassin, Cary Hudson, Casey Chu, Chad Nelson, Chak Li, Chan~Jun Shern, Channing Conger, Charlotte Barette, Chelsea Voss, Chen Ding, Cheng Lu, Chong Zhang, Chris Beaumont, Chris Hallacy, Chris Koch, Christian Gibson, Christina Kim, Christine Choi, Christine McLeavey, Christopher Hesse, Claudia Fischer, Clemens Winter, Coley Czarnecki, Colin Jarvis, Colin Wei, Constantin Koumouzelis, Dane Sherburn, Daniel Kappler, Daniel Levin, Daniel Levy, David Carr, David Farhi, David Mely, David Robinson, David Sasaki, Denny Jin, Dev Valladares, Dimitris Tsipras, Doug Li, Duc~Phong Nguyen, Duncan Findlay, Edede Oiwoh, Edmund Wong, Ehsan Asdar, Elizabeth Proehl, Elizabeth Yang, Eric Antonow, Eric Kramer, Eric Peterson, Eric Sigler, Eric Wallace, Eugene Brevdo, Evan Mays, Farzad Khorasani, Felipe~Petroski Such, Filippo Raso, Francis Zhang, Fred von Lohmann, Freddie Sulit, Gabriel Goh, Gene Oden, Geoff Salmon, Giulio Starace, Greg Brockman, Hadi
  Salman, Haiming Bao, Haitang Hu, Hannah Wong, Haoyu Wang, Heather Schmidt, Heather Whitney, Heewoo Jun, Hendrik Kirchner, Henrique~Ponde de~Oliveira~Pinto, Hongyu Ren, Huiwen Chang, Hyung~Won Chung, Ian Kivlichan, Ian O'Connell, Ian O'Connell, Ian Osband, Ian Silber, Ian Sohl, Ibrahim Okuyucu, Ikai Lan, Ilya Kostrikov, Ilya Sutskever, Ingmar Kanitscheider, Ishaan Gulrajani, Jacob Coxon, Jacob Menick, Jakub Pachocki, James Aung, James Betker, James Crooks, James Lennon, Jamie Kiros, Jan Leike, Jane Park, Jason Kwon, Jason Phang, Jason Teplitz, Jason Wei, Jason Wolfe, Jay Chen, Jeff Harris, Jenia Varavva, Jessica~Gan Lee, Jessica Shieh, Ji~Lin, Jiahui Yu, Jiayi Weng, Jie Tang, Jieqi Yu, Joanne Jang, Joaquin~Quinonero Candela, Joe Beutler, Joe Landers, Joel Parish, Johannes Heidecke, John Schulman, Jonathan Lachman, Jonathan McKay, Jonathan Uesato, Jonathan Ward, Jong~Wook Kim, Joost Huizinga, Jordan Sitkin, Jos Kraaijeveld, Josh Gross, Josh Kaplan, Josh Snyder, Joshua Achiam, Joy Jiao, Joyce Lee, Juntang
  Zhuang, Justyn Harriman, Kai Fricke, Kai Hayashi, Karan Singhal, Katy Shi, Kavin Karthik, Kayla Wood, Kendra Rimbach, Kenny Hsu, Kenny Nguyen, Keren Gu-Lemberg, Kevin Button, Kevin Liu, Kiel Howe, Krithika Muthukumar, Kyle Luther, Lama Ahmad, Larry Kai, Lauren Itow, Lauren Workman, Leher Pathak, Leo Chen, Li~Jing, Lia Guy, Liam Fedus, Liang Zhou, Lien Mamitsuka, Lilian Weng, Lindsay McCallum, Lindsey Held, Long Ouyang, Louis Feuvrier, Lu~Zhang, Lukas Kondraciuk, Lukasz Kaiser, Luke Hewitt, Luke Metz, Lyric Doshi, Mada Aflak, Maddie Simens, Madelaine Boyd, Madeleine Thompson, Marat Dukhan, Mark Chen, Mark Gray, Mark Hudnall, Marvin Zhang, Marwan Aljubeh, Mateusz Litwin, Matthew Zeng, Max Johnson, Maya Shetty, Mayank Gupta, Meghan Shah, Mehmet Yatbaz, Meng~Jia Yang, Mengchao Zhong, Mia Glaese, Mianna Chen, Michael Janner, Michael Lampe, Michael Petrov, Michael Wu, Michele Wang, Michelle Fradin, Michelle Pokrass, Miguel Castro, Miguel Oom~Temudo de~Castro, Mikhail Pavlov, Miles Brundage, Miles Wang, Minal
  Khan, Mira Murati, Mo~Bavarian, Molly Lin, Murat Yesildal, Nacho Soto, Natalia Gimelshein, Natalie Cone, Natalie Staudacher, Natalie Summers, Natan LaFontaine, Neil Chowdhury, Nick Ryder, Nick Stathas, Nick Turley, Nik Tezak, Niko Felix, Nithanth Kudige, Nitish Keskar, Noah Deutsch, Noel Bundick, Nora Puckett, Ofir Nachum, Ola Okelola, Oleg Boiko, Oleg Murk, Oliver Jaffe, Olivia Watkins, Olivier Godement, Owen Campbell-Moore, Patrick Chao, Paul McMillan, Pavel Belov, Peng Su, Peter Bak, Peter Bakkum, Peter Deng, Peter Dolan, Peter Hoeschele, Peter Welinder, Phil Tillet, Philip Pronin, Philippe Tillet, Prafulla Dhariwal, Qiming Yuan, Rachel Dias, Rachel Lim, Rahul Arora, Rajan Troll, Randall Lin, Rapha~Gontijo Lopes, Raul Puri, Reah Miyara, Reimar Leike, Renaud Gaubert, Reza Zamani, Ricky Wang, Rob Donnelly, Rob Honsby, Rocky Smith, Rohan Sahai, Rohit Ramchandani, Romain Huet, Rory Carmichael, Rowan Zellers, Roy Chen, Ruby Chen, Ruslan Nigmatullin, Ryan Cheu, Saachi Jain, Sam Altman, Sam Schoenholz, Sam
  Toizer, Samuel Miserendino, Sandhini Agarwal, Sara Culver, Scott Ethersmith, Scott Gray, Sean Grove, Sean Metzger, Shamez Hermani, Shantanu Jain, Shengjia Zhao, Sherwin Wu, Shino Jomoto, Shirong Wu, Shuaiqi, Xia, Sonia Phene, Spencer Papay, Srinivas Narayanan, Steve Coffey, Steve Lee, Stewart Hall, Suchir Balaji, Tal Broda, Tal Stramer, Tao Xu, Tarun Gogineni, Taya Christianson, Ted Sanders, Tejal Patwardhan, Thomas Cunninghman, Thomas Degry, Thomas Dimson, Thomas Raoux, Thomas Shadwell, Tianhao Zheng, Todd Underwood, Todor Markov, Toki Sherbakov, Tom Rubin, Tom Stasi, Tomer Kaftan, Tristan Heywood, Troy Peterson, Tyce Walters, Tyna Eloundou, Valerie Qi, Veit Moeller, Vinnie Monaco, Vishal Kuo, Vlad Fomenko, Wayne Chang, Weiyi Zheng, Wenda Zhou, Wesam Manassra, Will Sheu, Wojciech Zaremba, Yash Patil, Yilei Qian, Yongjik Kim, Youlong Cheng, Yu~Zhang, Yuchen He, Yuchen Zhang, Yujia Jin, Yunxing Dai, and Yury Malkov. 2024.
\newblock \href {https://arxiv.org/abs/2410.21276} {Gpt-4o system card}.
\newblock \emph{Preprint}, arXiv:2410.21276.

\bibitem[{Pourreza et~al.(2024{\natexlab{a}})Pourreza, Li, Sun, Chung, Talaei, Kakkar, Gan, Saberi, Ozcan, and Arik}]{CHASESQL}
Mohammadreza Pourreza, Hailong Li, Ruoxi Sun, Yeounoh Chung, Shayan Talaei, Gaurav~Tarlok Kakkar, Yu~Gan, Amin Saberi, Fatma Ozcan, and Sercan~O. Arik. 2024{\natexlab{a}}.
\newblock \href {https://arxiv.org/abs/2410.01943} {Chase-sql: Multi-path reasoning and preference optimized candidate selection in text-to-sql}.
\newblock \emph{Preprint}, arXiv:2410.01943.

\bibitem[{Pourreza and Rafiei(2023)}]{DINSQL}
Mohammadreza Pourreza and Davood Rafiei. 2023.
\newblock \href {https://arxiv.org/abs/2304.11015} {Din-sql: Decomposed in-context learning of text-to-sql with self-correction}.
\newblock \emph{Preprint}, arXiv:2304.11015.

\bibitem[{Pourreza and Rafiei(2024)}]{DTSSQL}
Mohammadreza Pourreza and Davood Rafiei. 2024.
\newblock Dts-sql: Decomposed text-to-sql with small large language models.
\newblock \emph{arXiv preprint arXiv:2402.01117}.

\bibitem[{Pourreza et~al.(2024{\natexlab{b}})Pourreza, Sun, Li, Miculicich, Pfister, and Arik}]{SQLGEN}
Mohammadreza Pourreza, Ruoxi Sun, Hailong Li, Lesly Miculicich, Tomas Pfister, and Sercan~O. Arik. 2024{\natexlab{b}}.
\newblock \href {https://arxiv.org/abs/2408.12733} {Sql-gen: Bridging the dialect gap for text-to-sql via synthetic data and model merging}.
\newblock \emph{Preprint}, arXiv:2408.12733.

\bibitem[{Rajkumar et~al.(2022)Rajkumar, Li, and Bahdanau}]{eval_llm}
Nitarshan Rajkumar, Raymond Li, and Dzmitry Bahdanau. 2022.
\newblock \href {https://arxiv.org/abs/2204.00498} {Evaluating the text-to-sql capabilities of large language models}.
\newblock \emph{Preprint}, arXiv:2204.00498.

\bibitem[{Shen et~al.(2025)Shen, Wan, Qiao, Zou, Xu, Shao, Zhang, Miao, and Pu}]{maplerepair}
Jiawei Shen, Chengcheng Wan, Ruoyi Qiao, Jiazhen Zou, Hang Xu, Yuchen Shao, Yueling Zhang, Weikai Miao, and Geguang Pu. 2025.
\newblock \href {https://arxiv.org/abs/2501.09310} {A study of in-context-learning-based text-to-sql errors}.
\newblock \emph{Preprint}, arXiv:2501.09310.

\bibitem[{Sun et~al.(2024)Sun, Arik, Muzio, Miculicich, Gundabathula, Yin, Dai, Nakhost, Sinha, Wang, and Pfister}]{SQLPaLM}
Ruoxi Sun, Sercan~Ö. Arik, Alex Muzio, Lesly Miculicich, Satya Gundabathula, Pengcheng Yin, Hanjun Dai, Hootan Nakhost, Rajarishi Sinha, Zifeng Wang, and Tomas Pfister. 2024.
\newblock \href {https://arxiv.org/abs/2306.00739} {Sql-palm: Improved large language model adaptation for text-to-sql (extended)}.
\newblock \emph{Preprint}, arXiv:2306.00739.

\bibitem[{Tai et~al.(2023)Tai, Chen, Zhang, Deng, and Sun}]{QDecomp}
Chang-You Tai, Ziru Chen, Tianshu Zhang, Xiang Deng, and Huan Sun. 2023.
\newblock \href {https://arxiv.org/abs/2305.14215} {Exploring chain-of-thought style prompting for text-to-sql}.
\newblock \emph{Preprint}, arXiv:2305.14215.

\bibitem[{Talaei et~al.(2024)Talaei, Pourreza, Chang, Mirhoseini, and Saberi}]{CHESS}
Shayan Talaei, Mohammadreza Pourreza, Yu-Chen Chang, Azalia Mirhoseini, and Amin Saberi. 2024.
\newblock \href {https://arxiv.org/abs/2405.16755} {Chess: Contextual harnessing for efficient sql synthesis}.
\newblock \emph{Preprint}, arXiv:2405.16755.

\bibitem[{Wang et~al.(2020)Wang, Shin, Liu, Polozov, and Richardson}]{RATSQL}
Bailin Wang, Richard Shin, Xiaodong Liu, Oleksandr Polozov, and Matthew Richardson. 2020.
\newblock {RAT-SQL}: Relation-aware schema encoding and linking for text-to-{SQL} parsers.
\newblock In \emph{Proceedings of the 58th Annual Meeting of the Association for Computational Linguistics}, pages 7567--7578, Online. Association for Computational Linguistics.

\bibitem[{Wang et~al.(2025)Wang, Ren, Yang, Liang, Bai, Chai, Yan, Zhang, Yin, Sun, and Li}]{MACSQL}
Bing Wang, Changyu Ren, Jian Yang, Xinnian Liang, Jiaqi Bai, LinZheng Chai, Zhao Yan, Qian-Wen Zhang, Di~Yin, Xing Sun, and Zhoujun Li. 2025.
\newblock \href {https://arxiv.org/abs/2312.11242} {Mac-sql: A multi-agent collaborative framework for text-to-sql}.
\newblock \emph{Preprint}, arXiv:2312.11242.

\bibitem[{Wei et~al.(2023)Wei, Wang, Schuurmans, Bosma, Ichter, Xia, Chi, Le, and Zhou}]{CoT}
Jason Wei, Xuezhi Wang, Dale Schuurmans, Maarten Bosma, Brian Ichter, Fei Xia, Ed~Chi, Quoc Le, and Denny Zhou. 2023.
\newblock \href {https://arxiv.org/abs/2201.11903} {Chain-of-thought prompting elicits reasoning in large language models}.
\newblock \emph{Preprint}, arXiv:2201.11903.

\bibitem[{Xie et~al.(2024)Xie, Wu, and Zhou}]{MAGSQL}
Wenxuan Xie, Gaochen Wu, and Bowen Zhou. 2024.
\newblock \href {https://arxiv.org/abs/2408.07930} {Mag-sql: Multi-agent generative approach with soft schema linking and iterative sub-sql refinement for text-to-sql}.
\newblock \emph{Preprint}, arXiv:2408.07930.

\bibitem[{Xu et~al.(2017)Xu, Liu, and Song}]{SQLNet}
Xiaojun Xu, Chang Liu, and Dawn Song. 2017.
\newblock \href {https://arxiv.org/abs/1711.04436} {Sqlnet: Generating structured queries from natural language without reinforcement learning}.
\newblock \emph{Preprint}, arXiv:1711.04436.

\bibitem[{Yu et~al.(2019)Yu, Zhang, Yang, Yasunaga, Wang, Li, Ma, Li, Yao, Roman, Zhang, and Radev}]{spider}
Tao Yu, Rui Zhang, Kai Yang, Michihiro Yasunaga, Dongxu Wang, Zifan Li, James Ma, Irene Li, Qingning Yao, Shanelle Roman, Zilin Zhang, and Dragomir Radev. 2019.
\newblock \href {https://arxiv.org/abs/1809.08887} {Spider: A large-scale human-labeled dataset for complex and cross-domain semantic parsing and text-to-sql task}.
\newblock \emph{Preprint}, arXiv:1809.08887.

\bibitem[{Zelle and Mooney(1996)}]{old_templates}
John~M. Zelle and Raymond~J. Mooney. 1996.
\newblock Learning to parse database queries using inductive logic programming.
\newblock In \emph{Proceedings of the Thirteenth National Conference on Artificial Intelligence - Volume 2}, AAAI'96, page 1050–1055. AAAI Press.

\bibitem[{Zhang et~al.(2024{\natexlab{a}})Zhang, Cao, Xu, Chen, and Yu}]{COESQL}
Hanchong Zhang, Ruisheng Cao, Hongshen Xu, Lu~Chen, and Kai Yu. 2024{\natexlab{a}}.
\newblock \href {https://arxiv.org/abs/2405.02712} {Coe-sql: In-context learning for multi-turn text-to-sql with chain-of-editions}.
\newblock \emph{Preprint}, arXiv:2405.02712.

\bibitem[{Zhang et~al.(2024{\natexlab{b}})Zhang, Chen, Liao, Wang, Zhao, Yu, Wang, Li, and Shi}]{SQLfuse}
Tingkai Zhang, Chaoyu Chen, Cong Liao, Jun Wang, Xudong Zhao, Hang Yu, Jianchao Wang, Jianguo Li, and Wenhui Shi. 2024{\natexlab{b}}.
\newblock \href {https://arxiv.org/abs/2407.14568} {Sqlfuse: Enhancing text-to-sql performance through comprehensive llm synergy}.
\newblock \emph{Preprint}, arXiv:2407.14568.

\end{thebibliography}
\clearpage
\newpage
\appendix
\section{Examples of Error Cause}
\begin{figure}[h]
    \centering
    \includegraphics[width=1.0\linewidth]{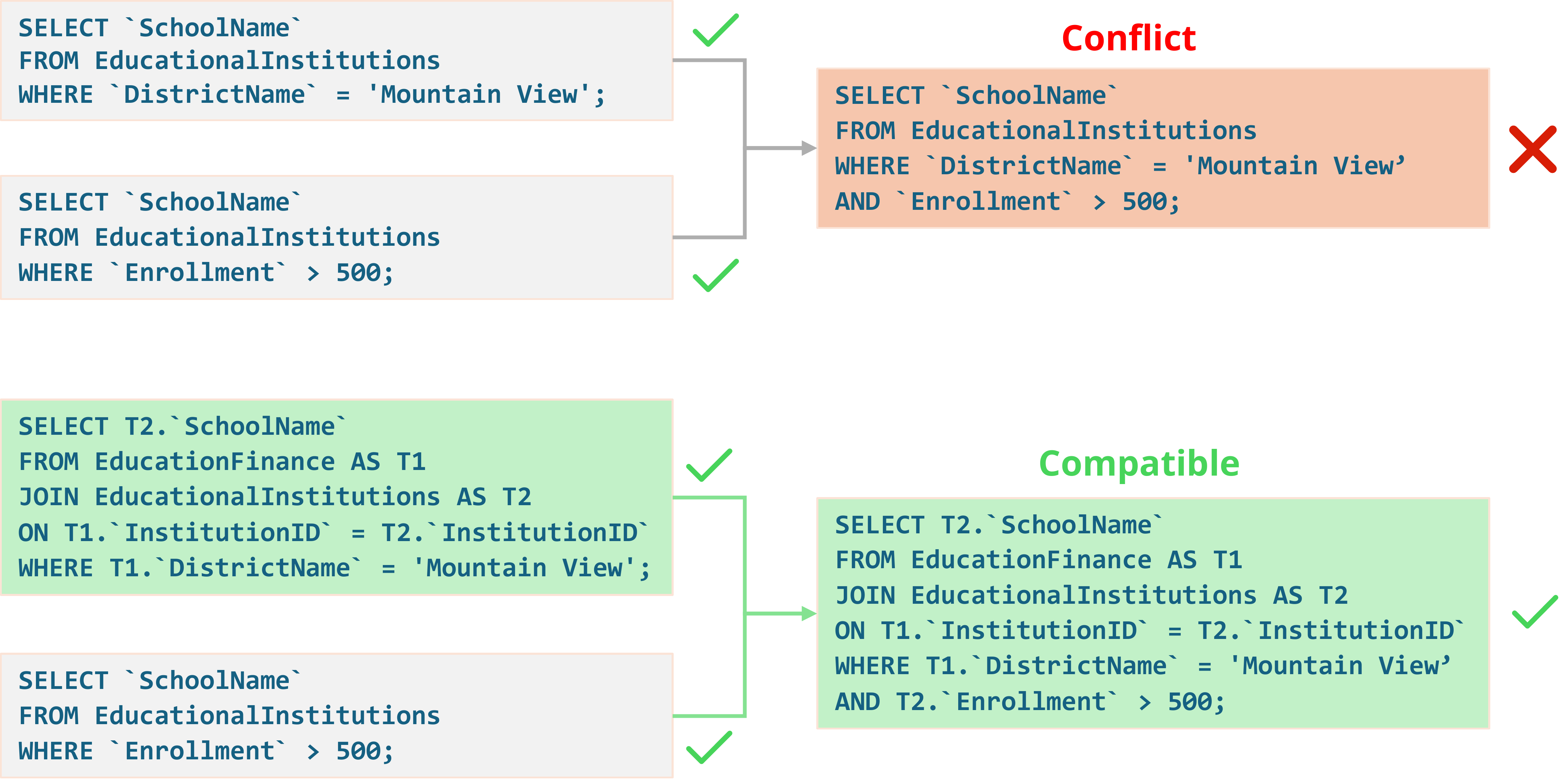}
    \caption{An Example of Condition Conflict}
    \label{fig:condition_conflict}
\end{figure}

\begin{figure}[h]
    \centering
    \includegraphics[width=1.0\linewidth]{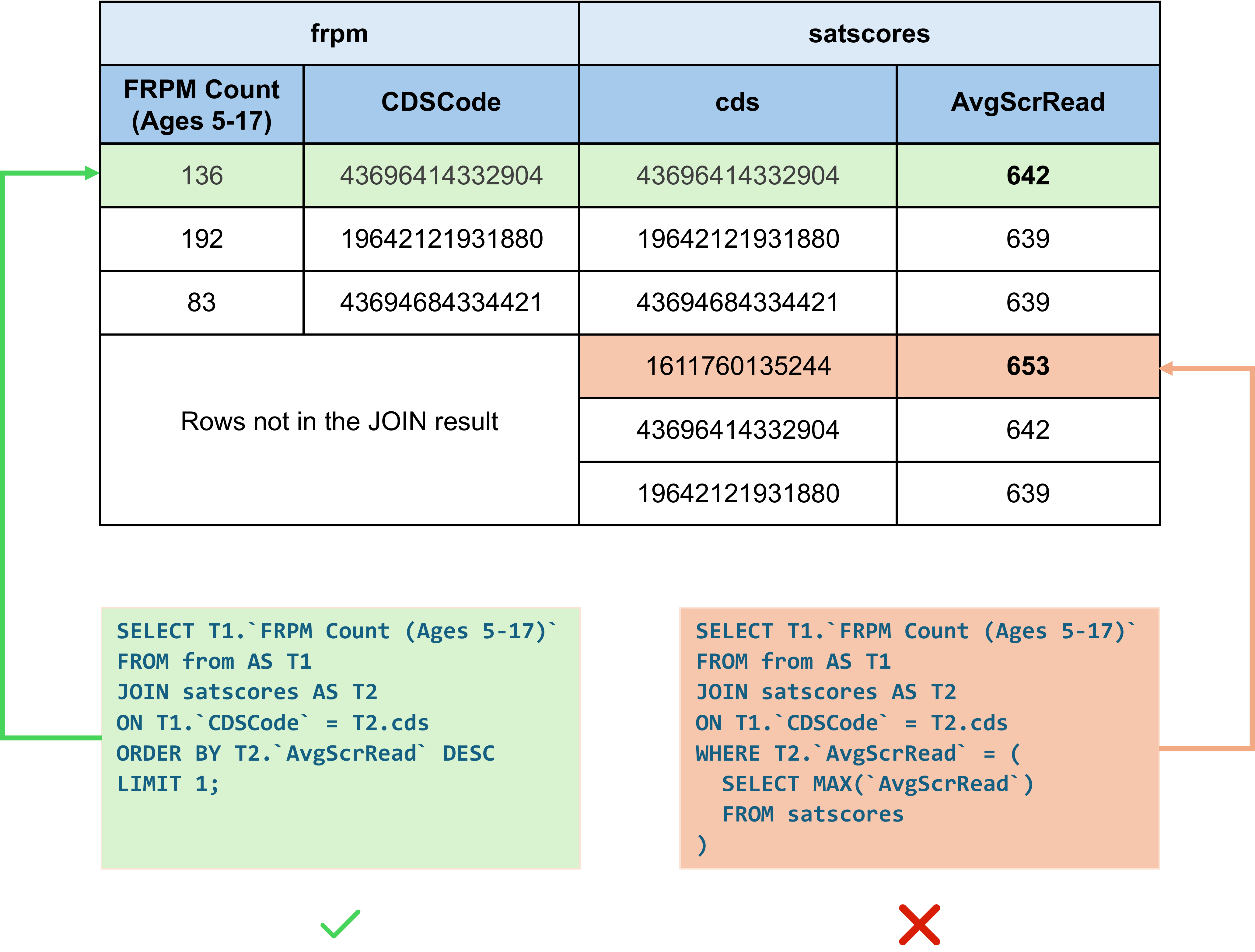}
    \caption{An example of Sub-query Scope Inconsistency}
    \label{fig:subquery_scope}
\end{figure}
\section{Target Checking Module}
\begin{figure}[h]
    \centering
    \includegraphics[width=\linewidth]{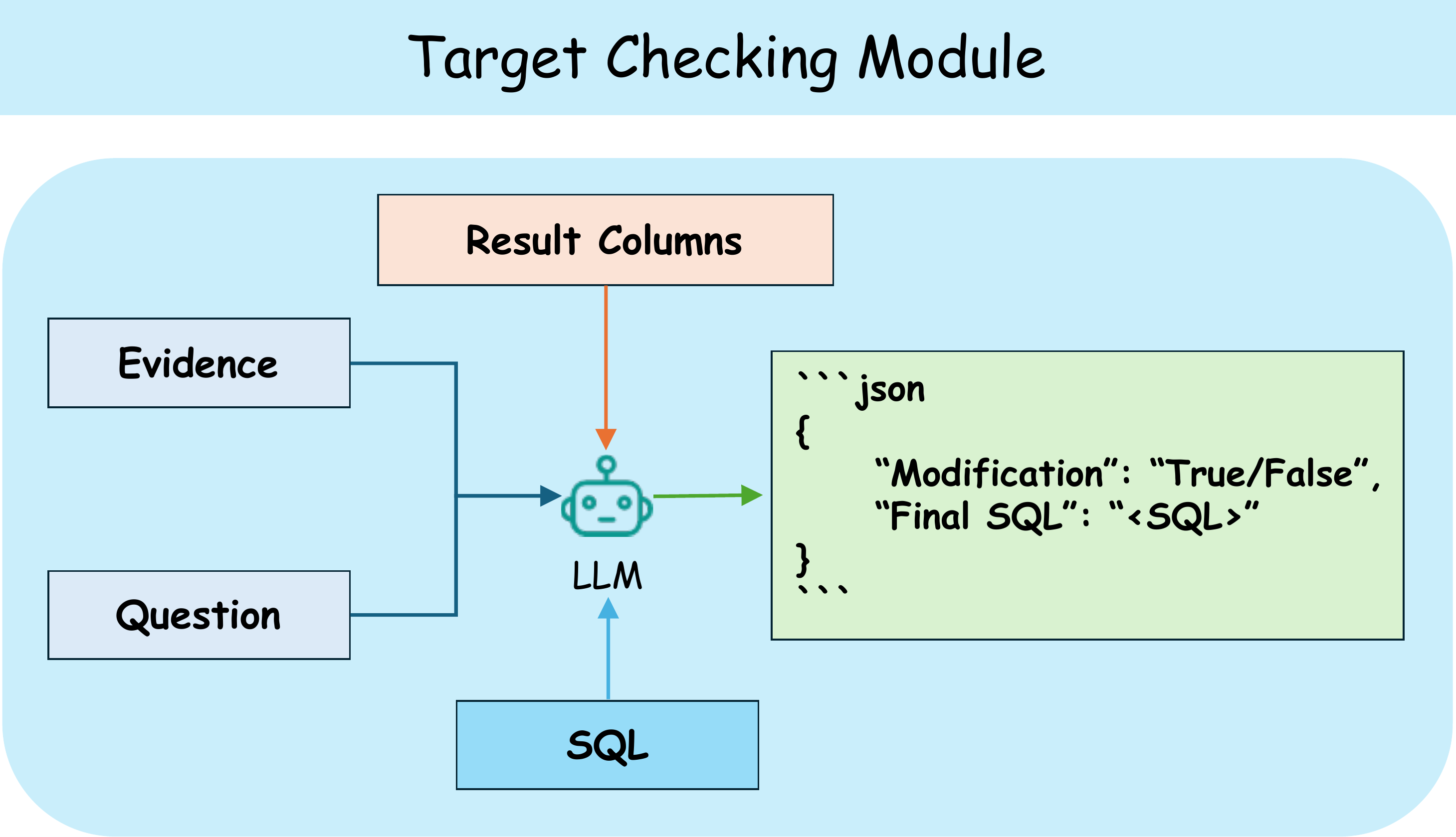}    \caption{Target Checking Module}
    \label{fig:target_check}
\end{figure}
\section{Training Settings}
\begin{table}[htbp]
\centering
\begin{tabular}{ll}
\toprule
\textbf{Parameter} & \textbf{Value} \\
\midrule
\texttt{per\_device\_train\_batch\_size} & 1 \\
\texttt{gradient\_accumulation\_steps} & 8 \\
\texttt{learning\_rate} & 1.0e-4 \\
\texttt{num\_train\_epochs} & 2.0 \\
\texttt{lr\_scheduler\_type} & cosine \\
\texttt{lora\_rank} & 16 \\
\bottomrule
\end{tabular}
\caption{Training hyperparameter configurations.}
\label{tab:train}
\end{table}
\newpage
\section{Figures}
\begin{figure}[h]
    \centering
    \includegraphics[width=1\linewidth]{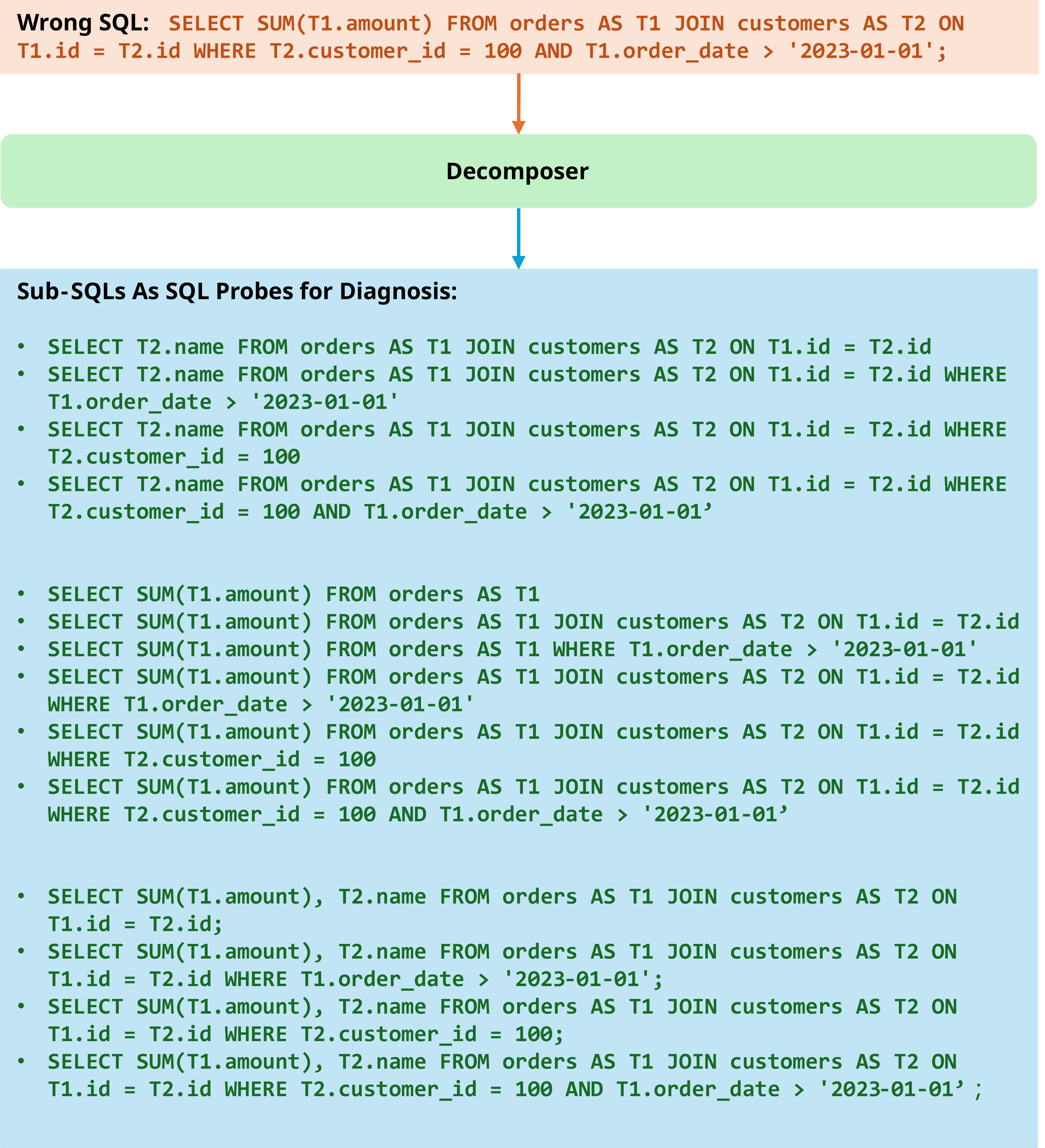}
    \caption{SQL Probes in Error Cause Identification Stage}
    \label{fig:error_search}
\end{figure}
\section{Prompt used by SDE-SQL in the training-free setting}
For approaches that do not rely on supervised training, it becomes particularly crucial to carefully craft and design prompt templates that can effectively guide the model to carry out Self-Driven Exploration behaviors in a controlled and meaningful manner. In this section, we present a comprehensive set of prompt templates that are utilized across different stages of the SDE-SQL pipeline to support this capability. Due to limitations in available space, certain detailed elements and specific prompt examples have been omitted, but the essential structures and core ideas are fully retained.

\begin{figure*}[h]
    \centering
    \includegraphics[width=1\linewidth]{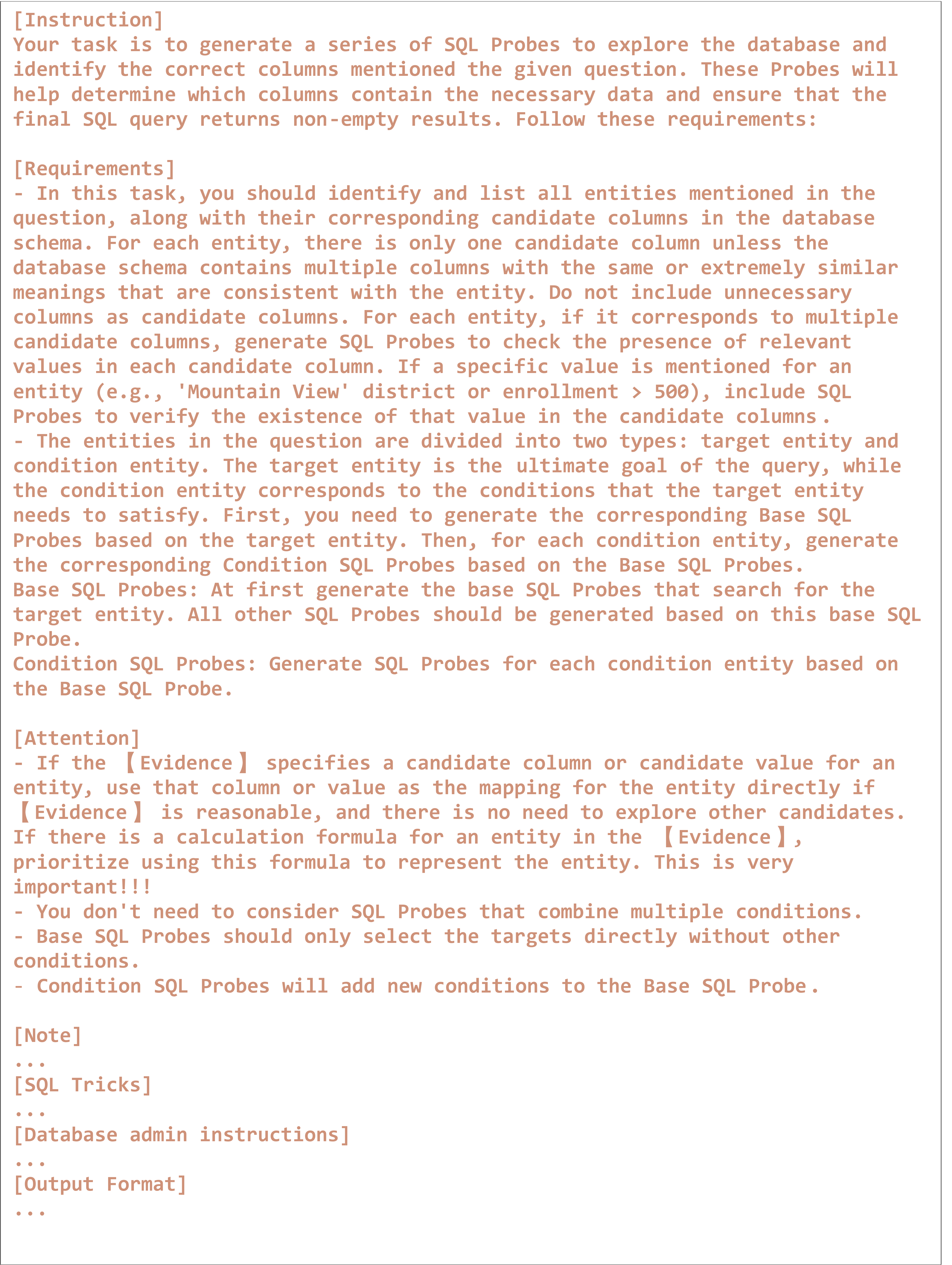}
    \caption{Prompt Template of Candidates Exploration}
    \label{fig:g1}
\end{figure*}

\begin{figure*}[h]
    \centering
    \includegraphics[width=1\linewidth]{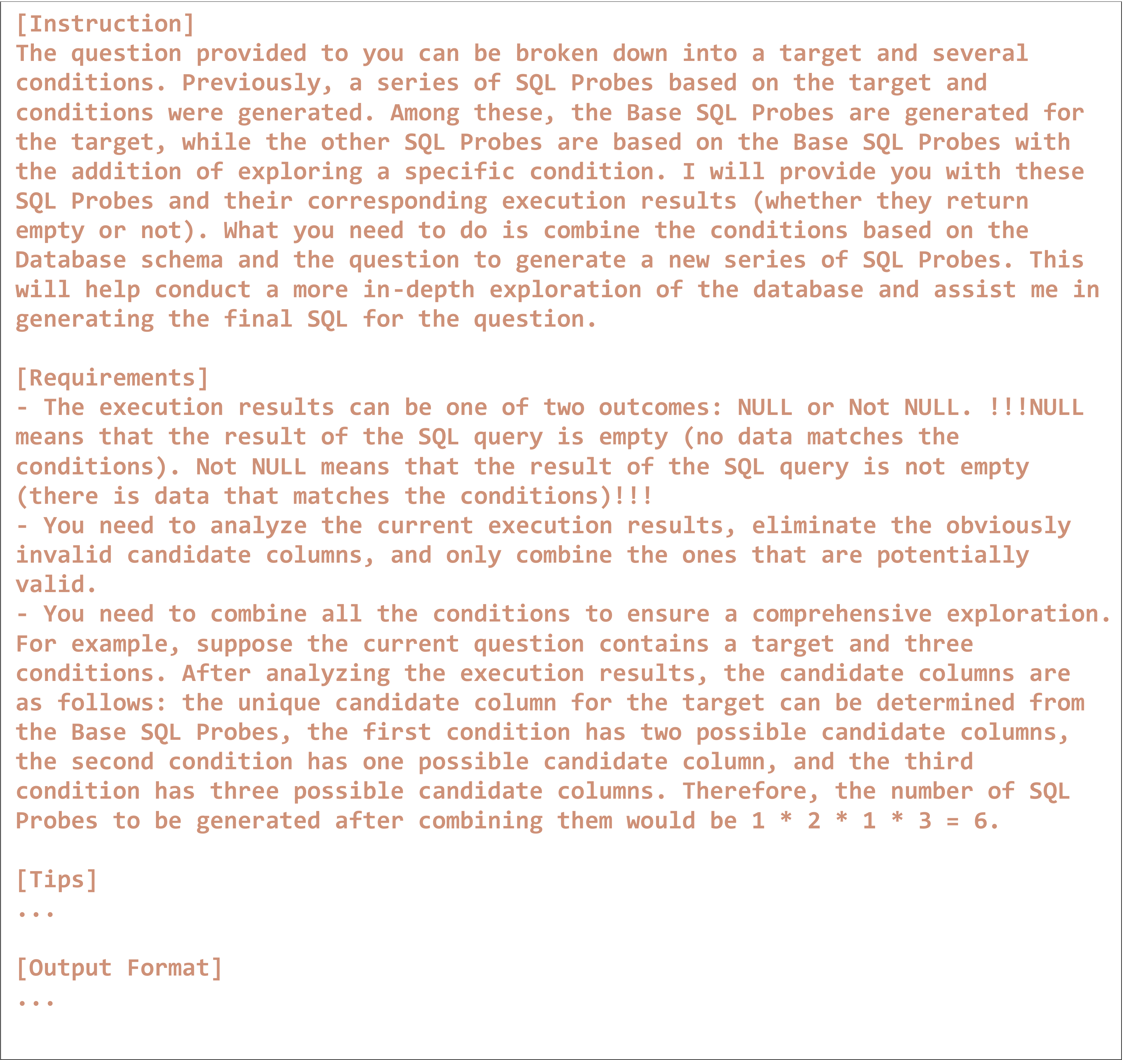}
    \caption{Prompt Template of Combinations Exploration}
    \label{fig:g2}
\end{figure*}

\begin{figure*}[h]
    \centering
    \includegraphics[width=1\linewidth]{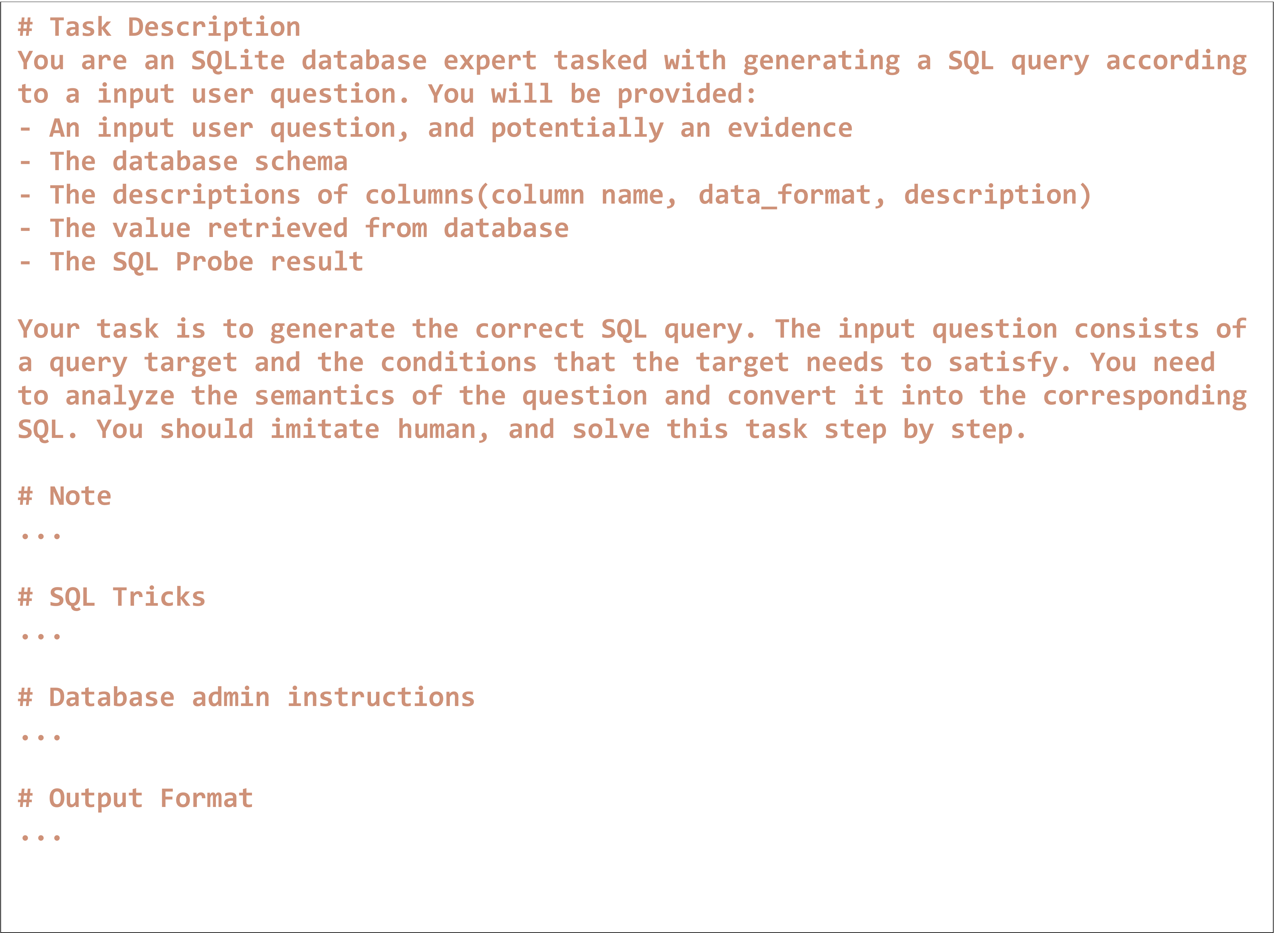}
    \caption{Prompt Template of Zero-shot Generation}
    \label{fig:g3}
\end{figure*}

\begin{figure*}[h]
    \centering
    \includegraphics[width=1\linewidth]{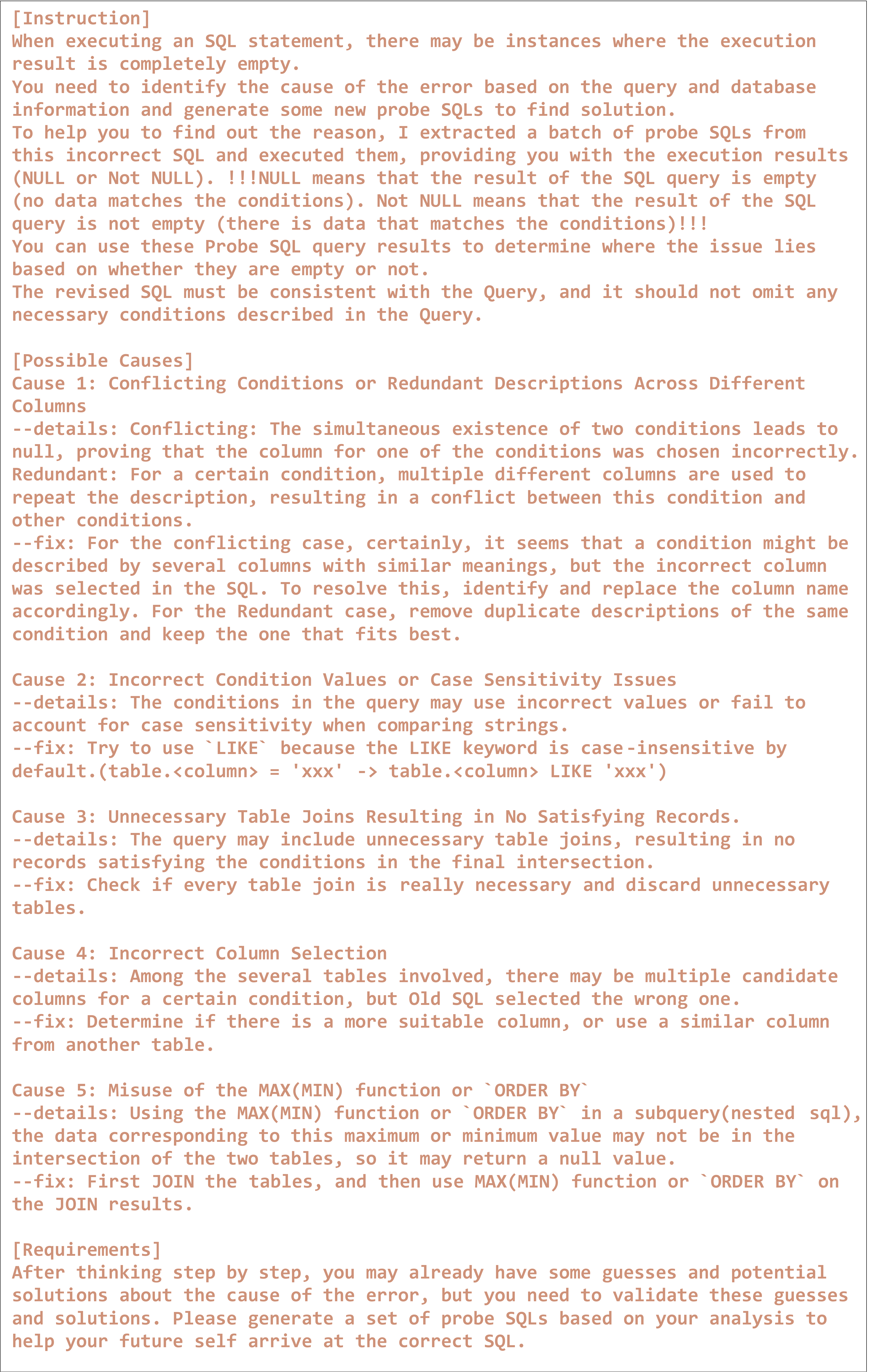}
    \caption{Prompt Template of Solution Exploration}
    \label{fig:g4}
\end{figure*}

\begin{figure*}[h]
    \centering
    \includegraphics[width=1\linewidth]{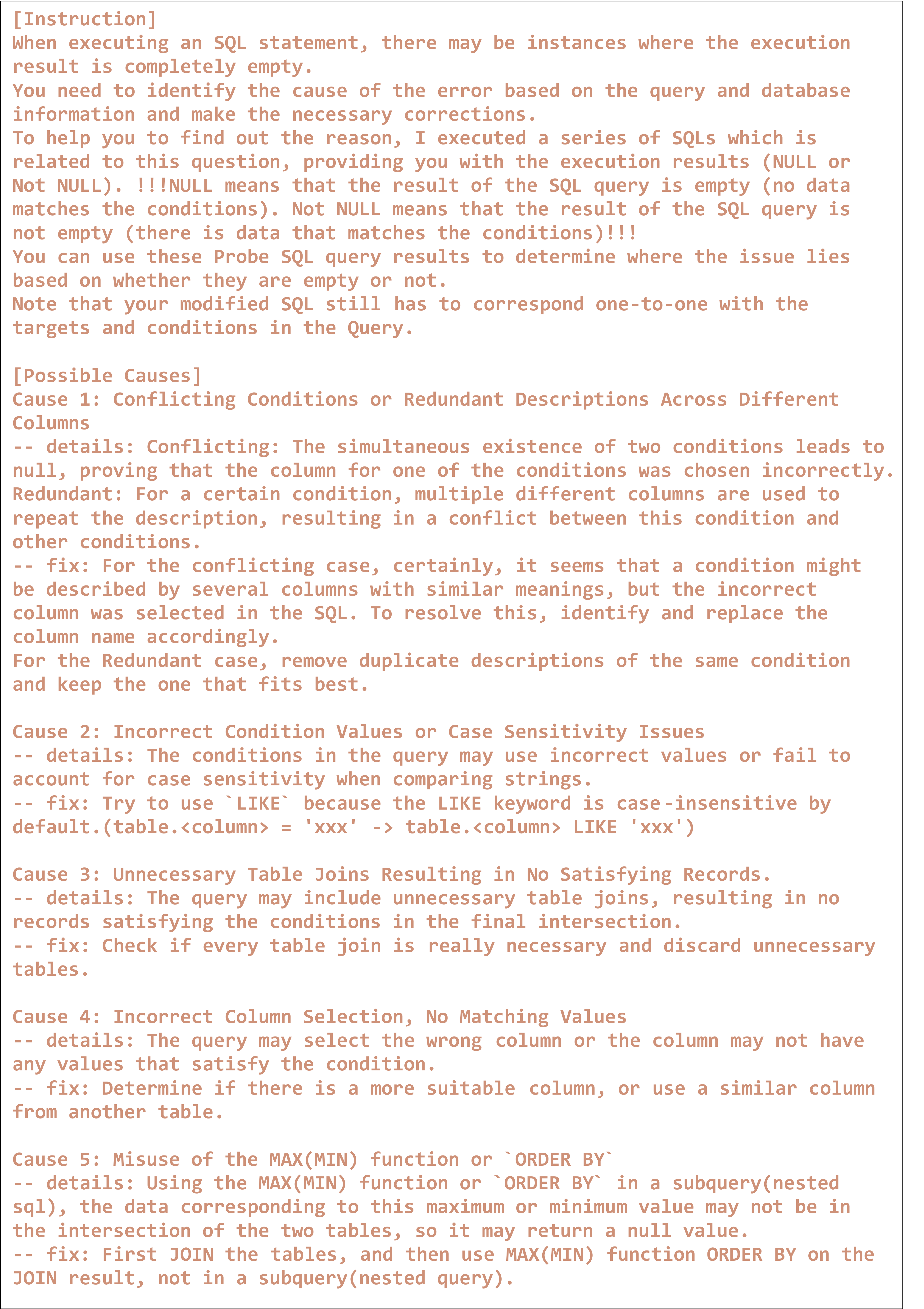}
    \caption{Prompt Template of Final Refinement}
    \label{fig:g5}
\end{figure*}

\begin{figure*}[h]
    \centering
    \includegraphics[width=1\linewidth]{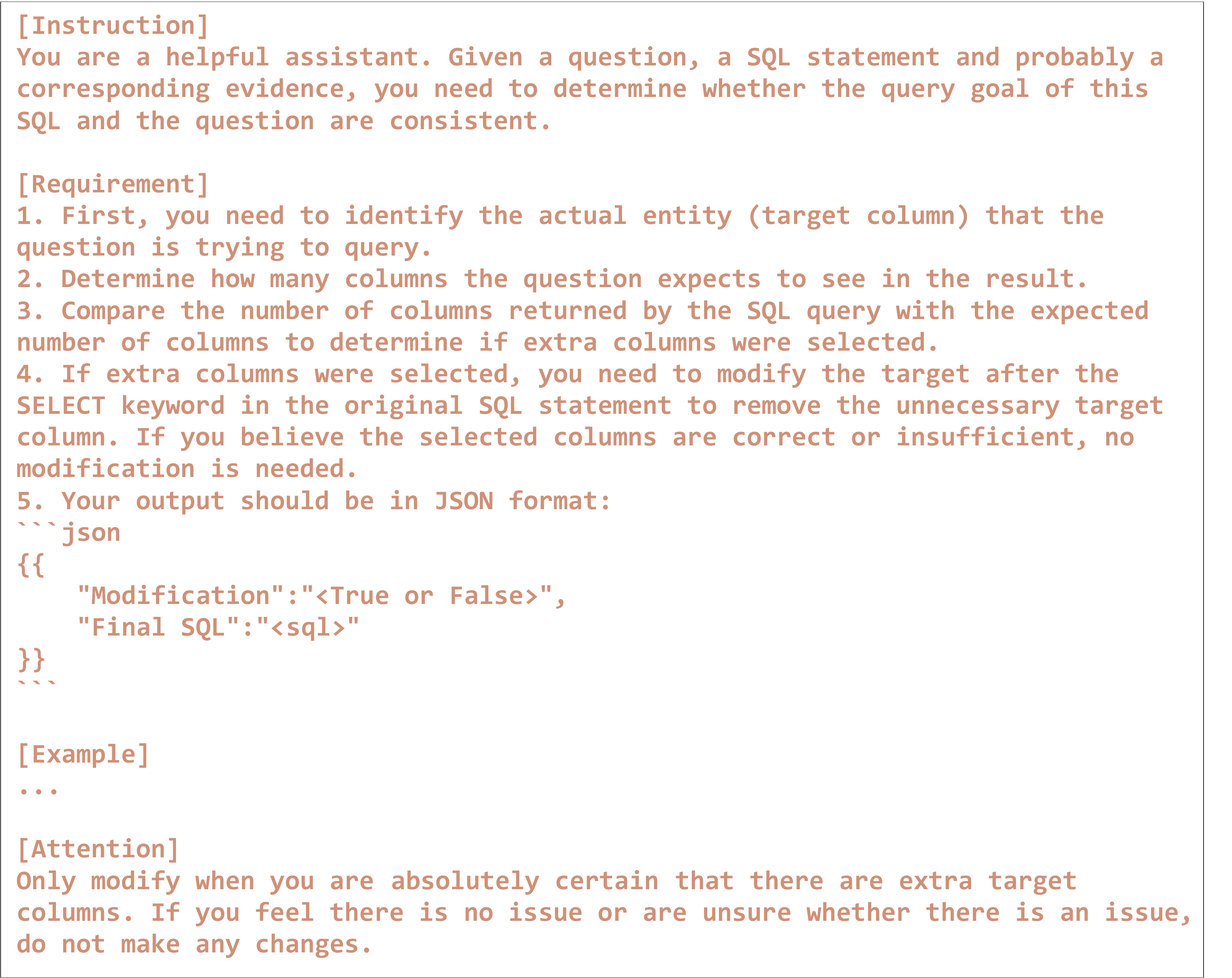}
    \caption{Prompt Template of Target Checking}
    \label{fig:g6}
\end{figure*}
\end{document}